\documentclass{article}

\PassOptionsToPackage{numbers}{natbib}
\usepackage[preprint]{neurips_2024}

\usepackage{graphicx}
\usepackage{multirow}
\usepackage{diagbox}
\usepackage{makecell}
\usepackage{wrapfig}
\usepackage{booktabs}
\usepackage{caption}
\usepackage{amsmath}

\usepackage[utf8]{inputenc} 
\usepackage[T1]{fontenc}    
\usepackage{hyperref}       
\usepackage{url}            
\usepackage{booktabs}       
\usepackage{amsfonts}       
\usepackage{nicefrac}       
\usepackage{microtype}      
\usepackage{xcolor}         
\usepackage{amsmath}
\usepackage{pifont}
\usepackage{enumerate}
\usepackage{enumitem}
\usepackage{bm}
\usepackage{ulem}
\usepackage{tabularx}

\usepackage{colortbl}
\usepackage{tcolorbox}
\definecolor{mygray}{gray}{.93}

\def\ours{TableGPT2{}}

\title{
\includegraphics[width=25pt]{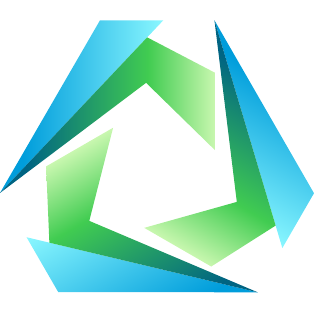}
\ours{}: A Large Multimodal Model \\ with Tabular Data Integration}

\author{
\parbox{0.90\linewidth}{Aofeng Su, Aowen Wang, Chao Ye, Chen Zhou, Ga Zhang, Gang Chen, Guangcheng Zhu, Haobo Wang, Haokai Xu, Hao Chen, Haoze Li, Haoxuan Lan, Jiaming Tian, Jing Yuan, Junbo Zhao\thanks{~Correspondence: \texttt{j.zhao@zju.edu.cn}.}, Junlin Zhou, Kaizhe Shou, Liangyu Zha, Lin Long, Liyao Li, Pengzuo Wu, Qi Zhang, Qingyi Huang, Saisai Yang, Tao Zhang, Wentao Ye, Wufang Zhu, Xiaomeng Hu, Xijun Gu, Xinjie Sun, Xiang Li,  Yuhang Yang, Zhiqing Xiao
}
\AND
Authors are ordered alphabetically by the first name.
\\
\makecell{
\\
Zhejiang University
\quad
Institute of Computing Innovation, Zhejiang University}
}
\vspace{-5pt}

\begin{document}

\maketitle

\vspace{-10pt}
\begin{abstract}

The emergence of models like GPTs, Claude, LLaMA, and Qwen has reshaped AI applications, presenting vast new opportunities across industries. Yet, the integration of tabular data remains notably underdeveloped, despite its foundational role in numerous real-world domains.

This gap is critical for three main reasons. First, database or data warehouse data integration is essential for advanced applications; second, the vast and largely untapped resource of tabular data offers immense potential for analysis; and third, the business intelligence domain specifically demands adaptable, precise solutions that many current LLMs may struggle to provide.

In response, we introduce \ours{}, a model rigorously pre-trained and fine-tuned with over \textbf{593.8K} tables and \textbf{2.36M} high quality query-table-output tuples, a scale of table-related data unprecedented in prior research. This extensive training enables \ours{} to excel in table-centric tasks while maintaining strong general language and coding abilities.

One of \ours{}’s key innovations is its novel table encoder, specifically designed to capture schema-level and cell-level information. This encoder strengthens the model’s ability to handle ambiguous queries, missing column names, and irregular tables commonly encountered in real-world applications. Similar to the VLMs, this pioneering approach integrates with the decoder to form a robust large multimodal model.

We believe the results are compelling: over \textbf{23 benchmarking metrics}, \ours{} achieves an average performance improvement of \textbf{35.20\%} in the 7B model and \textbf{49.32\%} in the 72B model over prior benchmark-neutral\footnote{~We focus exclusively on open-sourced benchmark-neutral LLMs, like Qwen~\cite{qwenteam2024qwen25}, LLaMA~\cite{dubey2024llama}, and DeepSeek~\cite{guo2024deepseek}, to ensure a fair, versatile comparison without overfitting to specific benchmarks. Specialized models like CHASE-SQL~\cite{pourreza2024chasesqlmultipathreasoningpreference} that are designed for specific tasks are generally excluded from our evaluation.} LLMs, with robust general-purpose capabilities intact.

We release an open-source repository (Section~\ref{sec:disclaimer}) that includes both the model and a comprehensive agent workflow, along with a subset of RealTabBench. This release aims to foster further exploration and application in real-world data-driven and BI production environments.

\end{abstract}

\clearpage
\tableofcontents
\clearpage

\section{Introduction}

The emergence of large language models (LLMs) has driven remarkable progress in artificial intelligence (AI), reshaping its applications across various domains. Models like ChatGPT~\cite{openai2023chatgpt} have enhanced the machine's ability to comprehend and produce human-like language, opening new possibilities in diverse fields.

\subsection{Motivation behind \ours{}}

\paragraph{A system perspective.}
\emph{No system works in a vacuum.}

The key motivation for proposing \ours{} is the need to address the limitations of current large language models (LLMs) in real-world, data-driven applications. No system operates in a vacuum, yet many LLMs are designed to function in an end-to-end fashion without external data integration. This approach is inherently flawed. For example, relying on an LLM to make stock investment recommendations without access to real-time market data is impractical. Similarly, an LLM making healthcare diagnoses without direct access to patient records or analyzing financial statements without the full dataset would lack the necessary depth and accuracy.

Even when LLMs are integrated with external data sources, such as databases, their performance is often suboptimal. One approach involves using tools like natural-language-to-sql (NL2SQL)~\cite{zhong2017seq2sql}, which provide schema metadata, table descriptions, and sample values to bridge the gap. However, as mentioned earlier, this method falls short in complex scenarios. Additionally, thanks to innovations like longer context windows or novel architectures, some models attempt to handle large datasets within a single context. This, too, is inefficient, as digesting detailed information from each cell of a large database or Excel file exceeds the capacity and utility of the extended context window, which was primarily designed for textual data.

Thus, there is a pressing need for new techniques that move beyond the vacuum operation paradigm. This is the motivation behind the development of \ours{}—a model designed to integrate and process tabular data directly and efficiently, overcoming the inherent limitations of current LLMs, especially towards production-level deployment.

\paragraph{A data perspective.}
A key factor in the success of large language models like GPT-4~\cite{achiam2023gpt} has been the availability of vast datasets sourced from the open web. GPT-4, for instance, was trained on hundreds of billions of tokens, drawn from a wide range of publicly available text data. This abundance of text allowed for effective scaling, enabling the model to capture complex linguistic patterns and perform well across various tasks.

Similarly, vision-language models (VLMs) have thrived by leveraging large-scale datasets that pair images with descriptive text~\cite{zhang2024vision}. These models, such as CLIP~\cite{radford2021learning} and DALL·E~\cite{ramesh2021zeroshottexttoimagegeneration}, succeed by aligning visual data with language, creating robust representations that allow them to excel in multimodal tasks. The richness and scale of both image and text data have been crucial to this success, demonstrating the power of large, diverse datasets.

Despite the significant focus on text and visual data, tabular data is equally abundant and critical across industries. It’s estimated that over 70\% of global data is in structured tabular form, whether stored in databases or spreadsheets. This vast resource points to the potential for developing a large-scale tabular model capable of harnessing the same scaling laws applied to text and images. By leveraging massive datasets of tabular data, along with schema metadata, we aim to explore whether these data formats can be modeled effectively, leading to functional and powerful models for business intelligence and other applications.

This approach aligns directly with our novel encoder design, which focuses on modeling the structure and content of tabular data. The tabular data encoder within \ours{} enables it to capture schema-level and cell-level information, opening the door for large-scale, data-driven performance improvements, similar to those observed in text and vision-based models.

\paragraph{A BI perspective.}

Business intelligence (BI), in particular, stands out as an area primed for innovation based upon LLMs. Traditional BI systems typically depend on fixed queries, static data structures, and inflexible interaction methods, which restrict their adaptability to dynamic business needs. LLMs offer a pathway to overcome these limitations by supporting more natural query interactions and accommodating a wide range of user intents.
 
Despite their promise however, LLMs still face significant challenges in being effectively integrated into BI applications. Issues such as computational efficiency~\cite{li2024personal}, the inability to fully comprehend tabular data~\cite{fang2024large}, and a lack of alignment with complex BI schema and user demands remain key hurdles. These limitations have prevented LLMs from delivering on their full potential in BI contexts, where precision, performance, and ease of use are paramount. 

Diving further, closest to our work, popular approaches such as NL2SQL~\cite{hu2023chatdb,zhong2017seq2sql,li2023graphix} and common table comprehension~\cite{jiang2022omnitab, ye2023large, zhang2024tablellama} have proven insufficient for modern BI systems due to several limitations. Table comprehension tasks often focus on small, clean datasets like those from Wikipedia, which fail to represent the complexity of real-world scenarios where tables are larger and more intricate. 
Similarly, NL2SQL models and benchmarks rely on well-structured queries and clean schema, but in actual BI environments, “dirty” schemas, missing values, and complex relational structures render these models ineffective. 
Furthermore, both approaches struggle to handle the diverse user intents, complex business logic, and multi-step reasoning required in real-world applications, highlighting their inability to scale to the demands of modern BI tasks.

\subsection{Upon the Previous Version of TableGPT}

In the initial version of TableGPT~\cite{zha2023tablegptunifyingtablesnature}, we introduced approaches such as structured Domain-Specific Languages (DSLs) and specialized table encoder to manage complex table-based queries. These techniques showed promise in structured data handling but also highlighted areas for further refinement. Our latest iteration, \ours{}, advances these foundations with substantial improvements. We have scaled up data and training protocols, carefully redesigned each component, and introduced additional technical enhancements to improve robustness, broaden applicability, and optimize performance in diverse BI tasks.

To provide context, \emph{\ours{} is a large-scale multimodal model available in two configurations—a 7-billion-parameter model and a 72-billion-parameter model—both based on the Qwen2.5 model family. Training encompassed over 86 billion tokens for continual pretraining (CPT), more than 437.5K table-language interleaved samples for encoder training and over 2.36M high quality query-table-output tuples being utilized for supervised fine-tuning. This scale of data is unprecedented in related research, ensuring that \ours{} meets the rigorous demands of modern applications involving structured or tabular data.}

\subsection{General Introduction of \ours{}}
\ours{}’s language framework is built on the Qwen2.5~\cite{qwenteam2024qwen25} architecture. It underwent continual pretraining (CPT), supervised fine-tuning (SFT), and an agent framework for production-level ability. These steps differ from traditional LLMs as our pretraining and fine-tuning place a stronger emphasis on coding, multi-turn reasoning, and tool usage. These characteristics ensure the model is not only adept at natural language processing but also highly capable of handling the intricate requirements of table-related tasks.

In addition, we preliminarily explored the possibilities of multimodal alignment for tabular data. Specifically, \ours{} innovatively incorporates a separate modality module specifically for reading and interpreting tabular data. Similar to vision-language models (VLMs), \ours{} involve a tabular data reading module that generates specialized embeddings concatenated with the token embeddings from textual input. This additional module allows \ours{} to better capture the structure and semantics of tabular data, enabling more accurate tabular comprehension in complex BI scenarios.

\begin{figure}[!t]
    \centering
    \includegraphics[width=\linewidth]{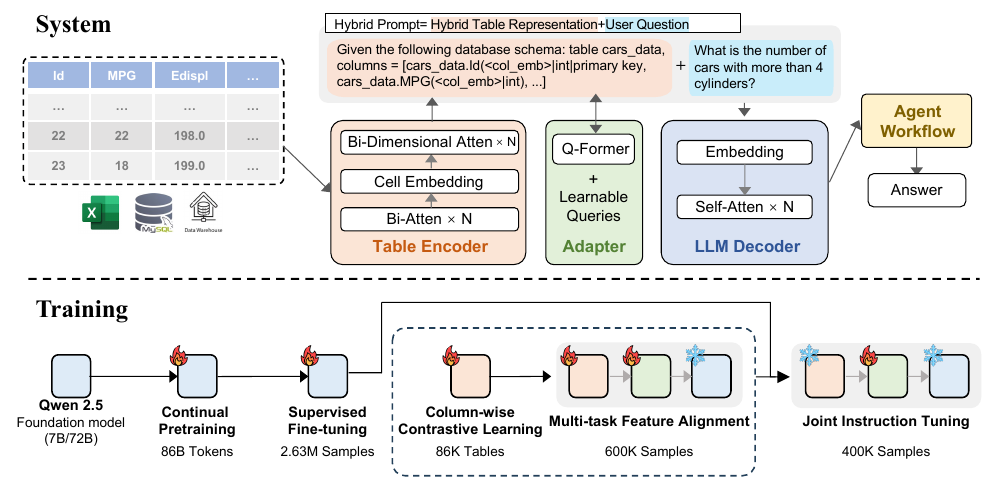}
    \caption{Overall framework of \ours{}.}
    \label{fig:main_system}
\end{figure}

An overall model framework is depicted in Figure~\ref{fig:main_system}.

\paragraph{Continual pretraining.}
For the goals of \ours{}, we first focused on enhancing the model’s coding and reasoning abilities through Continual Pretraining (CPT)~\cite{ke2023continual}. Specifically, 80\% of the CPT data was dedicated to well-commented code, aligning with approaches of DeepSeek-v2~\cite{liu2024deepseek}, ensuring robust coding capabilities. To complement this, we also curated substantial reasoning data and general textbooks containing domain-specific knowledge (e.g., finance, manufacturing, biotechnology, market-tech) to maintain a balanced data ratio for reasoning enhancement.

In terms of data processing, we employed a two-level filtering strategy. At the document level, we tagged the data using 54 distinct categories to ensure comprehensive coverage of different documentation types. Meanwhile, at the token level, we utilized the RHO-1~\citep{lin2024rho} technique to fine-tune the selection of high-quality tokens. Additionally, we introduced a novel approach to account for code length and context window settings~\cite{kim2024strategicdataorderingenhancing}, optimizing the model’s ability to handle various code segments effectively.

The final CPT data comprised 86B tokens, after thorough filtering, and this robust pretraining ensures that \ours{} is equipped with the necessary coding and reasoning capabilities tailored for complex BI and other related tasks.

\paragraph{Supervised fine-tuning.}
The purpose of Supervised Fine-Tuning (SFT) in \ours{} was to address the model’s limitations in adapting to BI-specific tasks and scenarios, where it lacked specialization. To improve its performance, we curated a dataset covering a wide range of critical and realistic scenarios, including multi-turn conversations, complex reasoning, tool usage, and highly business-specific queries.

The dataset was constructed using a combination of manual annotations and an expert-driven automatic labeling pipeline, ensuring the quality and relevance of the data. In total, the SFT process involved 2.36M samples, with a token count in the range of several billion, further refining the model to meet the specific demands of BI and other environments that involve tables.

More specifically, a key differentiator in \ours{}’s SFT process is the balanced and diverse composition of these 2.36M instructional samples. Given that table-related tasks require both general model capabilities and specialized table-specific skills, we ensured a well-organized mix of data. This dataset includes table-specific tasks such as code generation (Python and SQL), table querying, data visualization, statistical testing, and predictive modeling. Additionally, it spans a wide variety of tasks like table comprehension, table generation, missing value imputation, and table-based question-answering, covering almost all stages of table usage. The input formats combined with random arrangements of table metadata like field descriptions, schema information, and value enumeration, producing over 20 different combinations of table-info inputs to ensure comprehensive coverage.

To ensure the high quality of the data, we implemented a multi-step data filtering pipeline. First, a set of rule-based filters was applied, including checks for code executability and correctness using Python and SQL executors, which removed common errors such as key errors and type conversion issues. Regular expressions and additional rules were also used to discard anomalous outputs, such as outputs that improperly redefined the table or returned comments without executable code. The filtered data was then subject to scoring, using multiple models (e.g., GPT-4o~\cite{hurst2024gpt}) and tailored prompts for more granular evaluation. Only samples that exceeded a set threshold across all scoring combinations were retained. Following this, sample calibration was performed through manual checks. If the sample accuracy fell below 95\%, the data generation and filtering scripts were reviewed and optimized. Lastly, evaluation was conducted using a fixed validation set of approximately 94.9K cases (including existing and our constructed cases), ensuring the generated results were executable and accurate, with further manual verification to spot-check any inconsistencies and detect potential data issues, such as missing function calls or multi-turn dialogue capability gaps.

\paragraph{Data augmentation for tabular data.}

To enhance \ours{}’s performance, especially in handling complex BI-related tasks, we implemented several query augmentation techniques. By introducing fuzziness in how fields were referenced within queries, we minimized direct mappings between query terms and table names or fields. This approach helped reduce post-training key errors in the model. Additionally, we applied table data augmentation by anonymizing field names and category values, which improved the model’s robustness when dealing with noisy or incomplete datasets, making it more resilient to “dirty” table data often encountered in real-world applications.

Furthermore, we reinforced the model’s versatility by incorporating single-turn and multi-turn QA tasks, using varied prompt formats and output structures to reduce \ours{}’s sensitivity to specific prompt templates. To further diversify the training data, we applied post-processing enhancements during data generation. This involved randomly adding details like visualization instructions or summary explanations, ensuring that the generated data samples were more varied and capable of addressing a wide range of real-world scenarios.

\paragraph{Semantic table encoder.}
The design of the semantic encoder in \ours{} is motivated by the limitations of conventional workflows in tasks like NL2SQL, which typically rely on predefined schemas and a few (or completely no) example cell values. While schemas may provide structural information, they often lack the depth and specificity required to capture the true semantics of a table. These methods often fall short when broader, more comprehensive information from the table’s content is required to perform accurate coding or analysis. In many real-world cases, the context needed for proper reasoning spans a much larger portion of the table, making it impractical to rely solely on individual cell values or small sections of data.

Current LLMs also struggle in this regard. While context window sizes have increased, allowing for more information to be processed at once, the performance of these models declines significantly when faced with too many cell values or database-level data. The sheer volume of information exceeds the capacity of the model’s attention mechanisms, leading to a loss of context and accuracy in generating code or performing analysis. Moreover, tables possess unique properties such as bi-dimensional structure, redundancy, and sparsity, which creates a significant gap between understanding table data and ordinary text.

To address this, we designed a tabular encoder that takes the entire table as input, producing a set of compact embeddings for each column. This architecture is tailored to the unique properties of tabular data, which differ fundamentally from text, images, or other data types. Table semantics arise from four key dimensions: cells, rows, columns, and the entire table structure, all of which exhibit permutation invariance. To accommodate this, we implemented a bi-dimensional attention mechanism without positional embeddings, alongside a hierarchical feature extraction process. This ensures that both row-wise and column-wise relationships are captured and effectively understood. Additionally, a column-wise contrastive learning approach is employed, encouraging the model to learn meaningful, structure-aware semantic representations of the table.

Column embeddings are aligned with textual embeddings through a Q-former-style adapter~\cite{li2023blip}, which is equipped with a set of learnable queries. We also introduce two special tokens, "<tab>" and "</tab>", to differentiate tabular features from the native text, allowing the model to process both modalities concurrently and without confusion. To further enhance the alignment between the textual information, column embeddings, and schema metadata, we applied joint instruction tuning. This process helps refine the model’s understanding of the tabular data, enabling it to more effectively integrate and interpret the various inputs.

\paragraph{Agent framework.}
In our open-source repository, we provide a comprehensive agent workflow runtime framework designed to seamlessly integrate \ours{} with enterprise-level data analysis tools. This framework consists of three core components: runtime prompt engineering, a secure code sandbox, and an agent evaluation module, which together enhance the agent’s capabilities and reliability. The workflow supports complex data analysis tasks through modular steps—input normalization, agent execution with optional VLM support, and tool calls—that work in tandem to manage and monitor the agent’s performance. By incorporating retrieval-augmented generation (RAG)~\cite{lewis2020retrieval} for efficient context retrieval and a code sandbox for secure execution, this framework ensures \ours{} delivers accurate, contextually relevant insights in real-world problems.

\paragraph{Evaluation.} 
We conducted extensive evaluations across various widely-used tabular and general benchmarks, demonstrating \ours{}’s exceptional capabilities in table comprehension, processing, and reasoning, with an average improvement of \textbf{35.20\%} for the 7B model and \textbf{49.32\%} for the 72B model while retaining robust general-purpose performance. \emph{Noted, for a fair assessment, we compare \ours{} only with open-sourced benchmark-neutral models such as Qwen~\cite{qwenteam2024qwen25} and DeepSeek~\cite{guo2024deepseek}, ensuring balanced, versatile performance across tasks without overfitting to any single benchmark}.
We also introduce and partially release a new benchmark, RealTabBench, which emphasizes unconventional tables, anonymized fields, and complex queries, aligning more closely with realistic scenarios.

\section{Disclaimer of Open Source}
\label{sec:disclaimer}

Our open-source repository comprises two main components: 
(i) \emph{\textbf{a standalone decoder}} opened weight on Hugging Face\footnote{\url{https://huggingface.co/tablegpt/tablegpt}} that achieves superior performance on table- and BI-related tasks without compromising general coding and conversational capabilities, as discussed in Section~\ref{sec:exp}, 
and (ii) \emph{\textbf{a holistic agent workflow}}, detailed in Section~\ref{sec:agent}, accessible via a GitHub repository\footnote{\url{https://github.com/tablegpt/tablegpt-agent}}. Additionally, \emph{\textbf{a portion of RealTabBench data}} is included in the repository.

Based on this, we offer \emph{\textbf{a few disclaimers}}:
\begin{enumerate}
    \item The \ours{} decoder functions effectively as a standalone model, particularly in scenarios without ambiguous queries, missing column names, or irregular tables (as in most existing benchmarks).
    
    \item The encoder-decoder configuration is recommended for more complex, real-world production scenarios, as the encoder’s pretraining enhances table comprehension and robustness.
    
    \item The encoder model is currently under preparation due to engineering considerations, including its complete integration with DeepSpeed for distributed training and vLLM for efficient inference.
    
    \item For RealTabBench, only a subset of the dataset is publicly released, with the remainder reserved for future evaluation as part of a comprehensive public benchmark to rigorously assess LLM performance in realistic BI environments.
\end{enumerate}

\section{Continual Pretraining from Qwen}

We selected Qwen-2.5~\cite{qwenteam2024qwen25}, the latest generation of open-sourced LLMs with strong multilingual support, as the base model for \ours{}. 
To better align the model with the specific requirements of business intelligence, we initiated a continual pretraining (CPT) process.

Compared to general usage, LLM for BI, or database-involved applications, often requires a distinct set of skills, with particular emphasis on tasks like data comprehension, complex analysis, and code generation. Recognizing these needs, we incorporated additional data types such as college-level textbooks, data analysis materials, and codes into our CPT process. This domain-specific data was supplemented with general corpora to ensure that the model retained broad language capabilities while gaining a deeper understanding of the targeted tasks.

\subsection{Data Collection}
Given the immense importance of coding abilities for BI applications, our data curation process is divided into two main categories: coding data and general data.

For coding data, we sourced primarily from StackOverflow (stack-v2) and GitHub, two of the most comprehensive repositories of code examples and developer discussions. After gathering the data, we performed a deduplication process to eliminate overlapping code and ensure that the dataset remained diverse and non-redundant.
For general data, we selected textbooks from fields such as finance, mathematics, and biology, ensuring a well-rounded representation of domain-specific knowledge. In addition to these, we incorporated data analysis materials (from the source including Kaggle~\cite{kaggle}) to further refine the model’s understanding of data-driven decision-making processes.

Special emphasis was placed on selecting data that included numbers, code snippets, and tables, as these elements are critical in the BI space. It ensures that \ours{} can efficiently handle tasks that require reasoning across both textual and structured data formats.

\begin{table*}[ht]
\centering
     \renewcommand{\arraystretch}{1.3}
     \setlength{\tabcolsep}{10pt} 
    \begin{tabular}{lcc}
    \toprule[0.4mm]
    \textbf{Data Type} & \textbf{Data Ratio} & \textbf{Number of Tokens (B)} \\
    \midrule
    Python         & 51.2\%  & 44.0  \\
    SQL            & 12.8\%  & 11.0  \\
    Other codes    & 16.0\%  & 13.8  \\
    General text   & 20.0\%  & 17.2  \\
    \midrule
    \textbf{Total} & \textbf{100\%} & \textbf{86.0} \\
    \bottomrule[0.4mm]
    \end{tabular}
    \caption{Data distribution and total tokens used in the Continual Pretraining (CPT) process.}
    \label{table:cpt}
\end{table*}

\subsection{Data Curation}
We adopted a two-stage filtering flow, including document-level and token-level filtering to ensure the data quality of the CPT procedure. 

\paragraph{Doc-level filtering.}
The primary objective of doc-level filtering is to remove low-quality data and ensure that only high-quality samples are used in the pretraining process. For general data, we employed methods similar to those used in C4~\cite{dodge2021documentinglargewebtextcorpora} and RefineWeb~\cite{penedo2023refinedwebdatasetfalconllm}, focusing on removing irrelevant or poor-quality content. We omit the detailed steps for simplicity, but it is important to note that keyword filtering was also applied to exclude sensitive and toxic corpora.

For coding data, we implemented a more complex selection scheme. After applying methods similar to those used in StarCoder2~\cite{lozhkov2024starcoder2stackv2}, we tagged the coding data across 54 distinct categories (as shown in Table~\ref{table:tag}) to ensure that the data spans various programming domains and scenarios, covering a broad range of code quality and use cases. We utilized tools like FastText~\cite{joulin2016bagtricksefficienttext}, which were distributed across a Ray thread pool for efficient processing. The filtering process was finalized through a set of heuristics, informed by the computational results from these tools.

\begin{table*}[ht]
\centering
     \renewcommand{\arraystretch}{1.5}
    \begin{tabular}{lll}
    \toprule[0.3mm]
    \textbf{Tag Name} & \textbf{Tag Description}    & \textbf{Data Type}  \\
    \hline
    \textit{lines\_num\_words} &  The number of words in each line & general, code\\
    \textit{fertility}    &  Character to token ratio &  general, code\\
    \textit{repetation\_ratio} & Repetitive line ratio & general \\
    \textit{language\_score} & Score of the language identification model & general \\
    \textit{compile} & Is the code executable & code \\
    \textit{nl\_ratio} & The proportion of natural language in the code & code \\
    \toprule[0.3mm]
    \end{tabular}
    
    \caption{Important filtering tags for code data and general data.}
    \label{table:tag}
\end{table*}

The finalized (after de-duplication and filtering) version of data is depicted in Table~\ref{table:cpt}.

\paragraph{Token-level filtering.}
After completing the document-level filtering, we implemented token-level training data governance using the Selective Language Modeling (SLM) approach, inspired by Rho-1\cite{lin2024rho}, during continual pretraining. Unlike conventional language models that compute losses for all training tokens, we recognize that not all tokens in a corpus hold equal importance for model training. Ideally, we aim to prioritize tokens that lead to a greater reduction in loss during training.
To achieve this, we first selected a reference model trained on high-quality text. We then scored each token’s loss in the corpus using both the reference model and the model undergoing continual pretraining. Finally, we applied a threshold to filter out high-scoring tokens (i.e., those with a larger difference in loss) for training, retaining only the most valuable tokens.

In practice, we used Qwen2.5-7B-Instruct~\cite{qwenteam2024qwen25} as the reference model and performed selective pretraining across all data. With a threshold set at 0.6, each token was labeled, and tokens falling below this threshold were masked during training, excluding them from loss computation.

\subsection{Training Strategy}
We selected Qwen-2.5-Base~\cite{qwenteam2024qwen25} as the base model and trained with context lengths of 8192 tokens. To create batches of data, we shuffled and merged documents, then truncated them to the specified context lengths. A cosine learning rate schedule was employed, with a maximum learning rate of 3e-5 and 3\% warm-up steps. We used BFloat16 mixed precision to ensure training stability. 
To further reduce instances of truncation during data batching, we grouped the data based on token lengths, ensuring that each batch contained as many complete data samples as possible, minimizing the need for truncation and preserving the integrity of the data.

\section{Semantic Table Encoder}

\begin{figure}[!t]
    \centering
    \includegraphics[width=\linewidth]{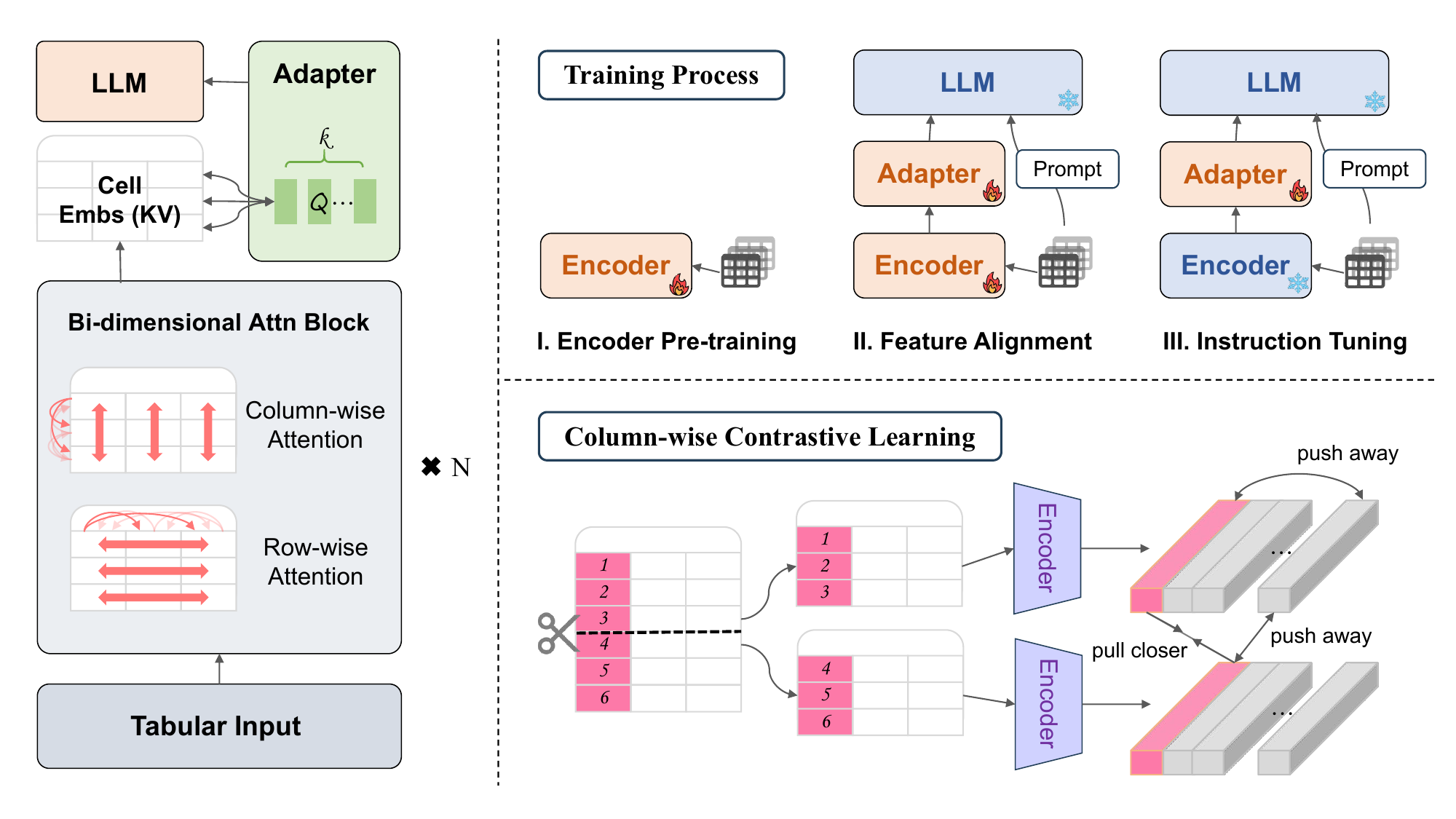}
    \caption{Overall design of the Semantic Table Encoder.}
    \label{fig:semenc}
\end{figure}

\subsection{Motivation}
The standard approach for utilizing LLMs in table-related tasks involves serializing tables into formats like Markdown, HTML, or XML to make them compatible with text-based inputs~\cite{li2024codes,pourreza2024din}. While larger context windows enable LLMs to handle more table content, a fundamental gap remains between language-based models and structured table understanding. This gap is evident in three key aspects: 1) \textbf{Structural incompatibility}: the bi-dimensional structure of tabular data is inherently misaligned with the sequential, autoregressive nature of language models, making LLMs less sensitive to structure-related patterns within the context. Previous studies have also highlighted this as a critical limitation of LLMs in table understanding~\cite{li2024unifying,yang2022tableformer}. 2) \textbf{Expressive inefficiency}: feeding LLMs with serialized tables is highly token-inefficient, particularly when dealing with large tables or databases. This leads to excessive computational costs and potential performance degradation due to the long contexts required. 3) \textbf{Lack of generalizability}: real-world tables are often noisy and ambiguous~\cite{yin2020tabert}, and when directly serialized into prompts, they can negatively impact the LLMs’ reasoning capabilities. Furthermore, due to context length limitations, including only a partial view of the table in the prompt may introduce context biases, misleading the model’s interpretation. To address these challenges, we propose a dedicated Semantic Table Encoder that provides the LLM with a structure-aware, token-efficient representation of table semantics.

\subsection{Model Architecture}
Our design for the Semantic Table Encoder consists of three main components: 1) the \textbf{semantic Table Encoder}, which generates structure-enriched semantic embeddings from the input tables; 2) the \textbf{table-language Adapter}, which aggregates the encoder’s output embeddings and aligns them with the textual features; and 3) the \textbf{backbone LLM}, which utilizes these high-level table representations to perform reasoning within the joint table-text space.

\paragraph{Table encoding.}
The design of the Semantic Table Encoder is tailored to the unique structural characteristics of tabular data. Given a table $\mathbf{T}=[\mathbf{c}_{11},\dots,\mathbf{c}_{1n};\dots,\mathbf{c}_{mn}]$ with $m$ rows and $n$ columns, where $\mathbf{c}_{ij}$ denotes the cell content in the $i$-th row and $j$-th column, we first apply a sentence transformer $\Phi$ to obtain compact semantic embeddings for each cell:
\begin{equation}
    \mathbf{E}(\mathbf{T})=[\Phi(\mathbf{c}_{11}),\dots,\Phi(\mathbf{c}_{mn})]\in \mathbb{R}^{m\times n\times d}
    \text{,}
\end{equation}
where $d$ is the dimension of each cell embedding.
These embeddings are then processed through a stack of bi-dimensional attention modules~\cite{somepalli2021saint,zhu2023xtab}, where they interact with other relevant cells to capture global structural table semantics:
\begin{equation}
    \mathbf{E}'(\mathbf{T})=\text{2D-Attn}(\mathbf{E}(\mathbf{T}))
    \in \mathbb{R}^{m\times n\times d}
    \text{.}
\end{equation}
Within each module, BERT-style bi-directional attention~\cite{kenton2019bert} is alternated along rows and columns, allowing the model to capture both intra-column distributional properties and inter-column relationships. To maintain the permutation invariance of (relational) tables, positional embeddings are intentionally excluded from these bi-dimensional attention modules.

\paragraph{Table-language adaption.}
To generate more compact table representations, we introduce an Adapter $g$ that aggregates cell-level embeddings at the column level and aligns them with the LLM’s text embeddings. The Adapter $g$ performs Q-Former-style cross-attention~\cite{li2023blip,bai2023qwen,wang2024qwen2} between $k$ learnable queries and the cell embeddings from each column, transforming tables with $m$ rows into fixed-length column representations of size $k$, matched to the LLM’s embedding dimension $d’$. The resulting embedding for the $i$-th column is computed as:
\begin{equation}
    \mathbf{C}(\mathbf{T})_i=g([\mathbf{E}'(\mathbf{T})_{1i}, \dots, \mathbf{E}'(\mathbf{T})_{mi}])
    \in \mathbb{R}^{k\times d'}
    \text{.}
\end{equation}
In this way, for an entire table, we obtain a compact representation $\mathbf{C}(\mathbf{T})\in\mathbb{R}^{n\times k\times d'}$ that encapsulates the global structure-aware table semantics and is directly compatible with the LLM for downstream tasks.

\definecolor{LBlue}{rgb}{0.76471, 0.92157, 0.980392}

\paragraph{Dynamic context integration.}
To seamlessly incorporate column embeddings with other textual table metadata (e.g., schemas and example cell values), we introduce a hybrid table representation that unifies these elements for efficient integration: 
\begin{center}
        {\centering \colorbox{LBlue}{{``table {\footnotesize\texttt{tab\_name}}, columns=[{\footnotesize\texttt{tab\_name}}.{\footnotesize\texttt{col\_name}}({\footnotesize\texttt{<col\_emb>}}\textbar{}{\footnotesize\texttt{dtype}}\textbar{}{\footnotesize\texttt{if\_primary\_key}})\textbar{}[\texttt{values}]]''}}.} 
\end{center}
During inference, the respective column embeddings are dynamically inserted into the specified slots at the LLM's embedding layer. This hybrid representation relieves the Table Encoder from losslessly compressing all table details, while still allowing them to provide the LLM with valuable high-level insights.

\subsection{Training}
\paragraph{Encoder pre-training.}
Similar to pre-training in other modalities, our objective in this stage is to equip the encoder with foundational tabular knowledge using a large corpus of raw tables. Inspired by how humans naturally interpret tables --- identifying patterns within columns and distinguishing differences between them --- we leverage contrastive learning to explicitly guide the Table Encoder in capturing intra-column semantics while differentiating across columns in a schema-independent manner. As shown in Figure~\ref{fig:semenc}, we first apply random row sampling (without replacement) on each table $\mathbf{T}_i$ within the mini-batch, creating two snapshots, $\mathbf{S}_i$ and $\mathbf{S}_i'$, which share the same schema but varying cell values. The Table Encoder is then used to generate an embedding pool $P$ consisting of column embeddings from each snapshot in the mini-batch. Then we perform contrastive learning, where positive pairs are formed by the embeddings from the same columns across the two snapshots. The contrastive loss~\cite{he2020momentum,chen2020simple} is defined as: 
\begin{equation}
    \mathcal{L}_{\text{cont}}(\tau, P) = - \frac{1}{|P|}\sum_{\bm{e}\in P} \log \frac{\exp(\bm{e}^\top \bm{e}_+ / \tau)}{\sum_{\bm{e}' \in P\setminus\{\bm{e}\}} \exp(\bm{e}^\top \bm{e}' / \tau)}\text{,}
\label{eqa:cont_loss}
\end{equation} 
where $\bm{e}_+$ is the embedding of the same column as $\bm{e}$ but from a different snapshot, and $\tau$ is the temperature. This approach encourages the model to learn distinct and context-sensitive column embeddings, capturing the unique semantics of each column within a given schema.

\paragraph{Feature alignment.}
To align tabular and textual features, we construct a set of multi-task table-language interleaved datasets and perform joint training of the Table Encoder and Adapter alongside a designated backbone LLM. Specifically, we create two synthetic interleaved datasets: \ding{172} \textbf{column prediction} --- predicting which column a given cell value belongs to, and \ding{173} \textbf{cell prediction} --- identifying which cell value corresponds to a given column, using pre-defined instruction templates and sampling questions and answers from raw tables without requiring manual annotations. Additionally, to further increase data diversity, we adapt three existing tabular datasets:  FetaQA~\cite{Nan2021FeTaQAFT}, WikiTableQuestion~\cite{wikitq}, and ToTTo~\cite{parikh-etal-2020-totto}, for augmented tasks: \ding{174} \textbf{question generation} --- generating a question based on a given answer from a specific table, \ding{175} \textbf{table titling} --- creating a brief title for a given table, and \ding{176} \textbf{row summarization} --- summarizing the content of a specific row. Each of these tasks is designed to maintain a high-level abstraction by minimizing dependency on specific cell values, promoting semantic table understanding. In total, we have collected 437K samples for feature alignment.

Tasks above all require the model to leverage the column embeddings to identify distributional patterns and inter-column relationships, facilitating a deeper integration and mutual alignment of the tabular and textual representations.

\paragraph{Joint instruction tuning.} 
To further enhance the model's instruction-following abilities for optimal performance on downstream tasks, we conduct joint supervised fine-tuning of the Semantic Table Encoder, Adapter, and LLM using a curated dataset for code generation. Relevant details will be elaborated on in Section~\ref{sec:sft_enc}.

\section{Supervised Fine-tuning over Tabular Data}
\subsection{Data Collection}

Recall that our aim is to handle flexible and dynamic data analysis tasks. We believe that supervision of processes involving the selection, analysis, computation, and manipulation of table contents is crucial. However, existing table analysis datasets fall short of providing such process-level supervision. A significant portion of TableQA datasets only contain table-question-answer pairs, which limits the activation of large models' capabilities for quantitative table operations. Meanwhile, available NL2SQL datasets are either too simplistic in question complexity or lack sufficient data volume. To address this gap, in addition to gathering existing tabular datasets, we have collected an additional batch of data containing 115K tables along with 479K samples. 
Particularly, our approach involves several key types of data: \emph{general corpora, coding corpora, and table-related query-answer pairs.}

In addition to these data types, our SFT process also incorporates both \textbf{single-turn} and \textbf{multi-turn} interactions, particularly for table-related tasks. This reflects real-world scenarios where multi-turn dialogue is often necessary to retrieve or analyze tabular data over multiple exchanges.

It is important to note that recent research~\cite{anonymous2024rethinking} has suggested that many Tabular LLMs fail at general language tasks as a trade-off for optimized table-related performance. However, in \ours{}, we maintain a balanced ratio of general data during fine-tuning. By carefully tuning the SFT process, we address this trade-off and ensure strong performance across both general and table-specific tasks, shown in Section \ref{sec:exp}. In the sequel, we illustrate our data collection process in detail. 

\subsubsection{Table Collection}

The tables to be collected are central to improving the performance of \ours{}. In general, the tables we collect cover a variety of domains, including but not limited to public government files, academics, manufacturing, finance, education, and healthcare.

The collected tables involve both regular and irregular structures. Regular tables are generally organized in a way that can be stored in databases or read into frameworks like pandas dataFrame. In contrast, irregular tables often contain merged cells or exhibit hierarchical structures. These irregular tables are often found in Excel files and tend to be smaller, with fewer columns and rows. Below is a summary of the types of table sources, their features, and relevant use cases:

\begin{itemize}
    \item \textbf{database tables}: These tables typically come from databases and involve multi-table fields, rows, columns, and large numeric data. These are often used in business scenarios involving frequent data queries. Example sources include public datasets such as MySQL's employee dataset, PostgreSQL's Chinook, and DBI documents.
    
    \item \textbf{web page tables}: These tables mainly record information within a single web page, such as data from official websites or academic journals. They are usually simple in form and include relevant contextual data.
    
    \item \textbf{excel tables}: These standalone files (e.g., .xls, .xlsx, .csv) contain various structured and unstructured information. These tables are often sourced from government data, financial reports, and business reports.
    
    \item \textbf{academic task tables}: These tables contain data used in research, often in structured forms suitable for TableQA or NL2SQL tasks. Sources include public datasets like WikiTableQuestions and other academic studies.
    
    \item \textbf{special format tables}: These tables have specific formats and usages, such as invoices, bills, and receipts, often used in online reporting systems and specialized software-generated documents.
    
    \item \textbf{pre-test task tables}: These tables are used for forecasting, prediction tasks, or scientific projects. Sources include Kaggle~\cite{kaggle}, UCI Machine Learning Repository~\cite{uci}, and Tianchi datasets~\cite{tianchi}.
\end{itemize}

The collection processes generally gather more than 160 sources, and the number of tables we collected varies from source to source.

\subsubsection{Query Collection Based on Tables}

Generating meaningful queries based on tables is a non-trivial task, though it is crucial for effective supervised fine-tuning of \ours{}. The goal is to provide a human or machine labeler with a table and prompt them to ask relevant and meaningful questions. These queries should not only reflect the data in the table but also be diverse in nature, ensuring they cover different types of instructions and are non-repetitive.

\paragraph{8 classes of queries.} 
To ensure comprehensive coverage of most Business Intelligence questions, we categorize them into 8 different categories: retrieval, insertion, deletion, modification, computation, statistics, visualization, and fuzzy queries. 

The last type, \textit{fuzzy queries}, targets questions that cannot be fully answered in a single conversation round. In such scenarios, \ours{} is designed to respond with follow-up questions to gather additional necessary information. 
For example, if a user asks: \textit{"What are the top-performing products?"}, \ours{} may not have enough context about which metric (e.g., sales, ratings, or profit margin) the user considers as a performance indicator. In this case, the model would respond with a clarifying question like: \textit{"Would you like to define performance by sales, profit, or customer reviews?"}

\paragraph{Synthesize and revise.} 

In our query generation process, we begin by synthesizing \textit{base} queries using the most advanced large language models (LLMs) with the largest parameter sizes. These base queries are then subjected to human revision to ensure quality and relevance.

For the synthesis phase, we employ models such as GPT-4o~\cite{hurst2024gpt}, LLaMa~\cite{dubey2024llama}, ChatGLM~\cite{glm2024chatglm}. Given a table, each of these LLMs generates 2-3 questions for each of categories. This results in a diverse set of base queries that form the foundation for further human refinement.

To optimize query generation, we explored various prompting schemes. We found that the prompt design needed to be customized for each LLM and query category. 
Additionally, we utilized \textbf{domain-specific jargon dictionaries} in combination with table-domain classifications to enhance the LLM's query generation. For instance, incorporating industry-specific terminologies significantly improved the relevance of queries. Retrieval-augmented generation (RAG)~\cite{lewis2020retrieval} plays a critical role in ensuring these specialized queries are accurate and aligned with the domain knowledge.

\paragraph{Multi-turn query generation.}

In this process, we address two key scenarios:

\begin{itemize}
    \item \textbf{scenario 1, reflection and refinement:} we aim to drive \ours{} to self-reflect and improve upon any improper answers. For this scenario, we use a weaker LLM to generate the first round of chain-of-thought (CoT) reasoning and code, followed by a stronger LLM for correction and refinement in the second round.

    \item \textbf{scenario 2, consecutive user queries:} in cases where users may ask a series of related questions about a given table, we generate context-dependent question pairs. This involves training the LLM to successfully generate multi-turn conversational queries that depend on previous interactions. For example: 'How many purchase invoices were recorded on February 1, 2022?', 'What is the total amount of these purchase invoices?' then 'What is the average tax amount for these purchase invoices?'.
\end{itemize}

All the generated queries, from both scenarios, are subsequently reviewed and refined by human labelers to ensure high quality and decent deviersity.

More specifically, for multi-turn query generation, the prompts span across various domains including manufacturing, finance, retail, healthcare, education, and other common industries (leveraging role-playing prompts). The multi-turn queries are categorized into 2-3 rounds, with approximately 100K data samples retained for each round, resulting in a total of around \textbf{160K multi-turn query sets}. For each table, we generated three sets of multi-turn queries, with each set containing 2-3 related questions. Considering that some cases involve multiple tables, the data covers approximately 120K tables.

However, noted, not all query sets correspond to complete multi-turn QA pairs. There are two cases where this happens:
\begin{itemize}
    \item Some data is filtered out during quality screening.
    \item If an error occurs during query generation or execution, preventing the completion of the query sequence, we truncate the queries accordingly. This means that, for some query sets consisting of 2-3 related questions, the final dataset may only retain a single query-answer sample.
\end{itemize}

\subsubsection{Answer Generation}

Given a provided table and a corresponding query, the main objective is to derive an accurate and relevant answer. Our approach follows a \emph{synthesize-then-refine} paradigm to ensure high-quality responses. We focus on two important aspects: single-turn and multi-turn answer generation. 

\paragraph{Single-turn answer generation.} As a simpler process, a single-turn answer does not require reflection on code execution errors, making it a good starting point for basic query handling. The aim of synthetic single-turn answer generation is to produce query-answer pairs where the code in the answers is executable and, ideally, as accurate as possible.

To achieve this, we use over 10 different prompt templates and, for each selected table and generated query set, randomly combine a template with the table information and a random query. One example template is:

\begin{quote}
\texttt{With several pandas dataframes available, your task is to write the Python code to address the user's question.}

\texttt{\#\# Follow this format:} \\
\texttt{Question: The user's query.} \\
\texttt{Thought: Evaluate the dataframes and the question to determine the solution.} \\
\texttt{Python code: Generate the Python code, within \textbackslash \textbackslash ```python ... \textbackslash \textbackslash ```}. \\

\texttt{\#\# Details about the dataframes:}

\texttt{\{df\_info\}}

\texttt{Question: \{input\}}
\end{quote}

We utilize advanced models such as GPT-4o~\cite{hurst2024gpt}, LLaMA~\cite{dubey2024llama}, ChatGLM~\cite{glm2024chatglm} to generate initial responses. The quality of these responses is first assessed by evaluating the execution of the generated Python code, as many code snippets fail to run correctly. After this initial check, the responses are reviewed and refined by human labelers. Both a preprocessing and postprocessing module are implemented. Preprocessing focuses on cleaning and standardizing table fields and values, ensuring that messy data fields are properly normalized. Postprocessing primarily involves filtering out code that fails to execute correctly, ensuring only functional outputs are retained. Human annotators serve as the final checkpoint, discarding any answers containing non-executable code or improper token generation.

\paragraph{Multi-turn answer generation.}
In our framework, we handle two distinct scenarios for multi-turn query generation. The first type, focusing on reflection and refinement, involves letting a weaker LLM generate code snippets, a textual answer and a CoT-styled plan. This initial answer is then passed to a more advanced, larger LLM to modify and refine. This is particularly valuable for distilling the ability to reflect on the correctness of the answers, reject faulty outputs, and iteratively improve the response.

The second scenario revolves around context-dependent queries, where each subsequent question depends on the answer provided in the previous turn. For example, in a scenario where a user asks, “What are the purchase invoice records for February 1, 2022?” and then follows up with, “What is the total amount of these invoices?” the LLM must handle the conversational flow and maintain context across turns. To achieve this, we adopt a ReAct prompting format (similar to the single-turn answer generation), allowing the LLM to interact with itself and produce a coherent, context-aware answer based on the ongoing query-answer dialogue.

For both of these scenarios, we initiate a code sandbox to observe and verify the coding answers. This sandbox plays a critical role in ensuring the executability of the code and the correctness of the output.

In general, we set the interaction to span 2-3 rounds of queries per conversation. However, due to the possibility of coding errors or non-executable code, the LLM is tasked with fixing these issues, which can extend the conversation length.
Finally, human labelers are responsible for verifying the generated answers. They ensure the answers are executable, accurate, and coherent across multiple turns, serving as the final stage in the process.

\subsubsection{Overall SFT Data Composition}
We iterated through our SFT (Supervised Fine-Tuning) data collection and refinement for 18 rounds, adjusting it in tandem with our training cycles. The final version of the SFT data composition reflects a comprehensive mixture of various query types, agent calls, programming languages, and table tasks. The overall data volume amounts to approximately \textbf{2.36M} query-answer samples.

Overall, our SFT dataset includes table-specific tasks such as code generation (Python and SQL), table querying, data visualization, statistical testing, and predictive modeling. Additionally, it spans a wide variety of tasks like table comprehension, table generation, missing value imputation, and table-based question-answering, covering almost all stages of table usage. The input formats were combined with random arrangements of table metadata like field descriptions, schema information, and value enumeration, producing over 20 different combinations of table-info inputs to ensure comprehensive coverage.

The distribution of different task categories is shown below:

\begin{table*}[htb]
  \centering
  \resizebox{0.9\linewidth}{!}{
    \begin{tabular}{l|c}
    \toprule
    \textbf{Category} & \textbf{Percentage} \\
    \midrule
    Agent \& Function Call & 2.09\% \\
    Coding Data Related to BI (Python \& SQL) & 21.51\% \\
    Table-Related Tasks (Single-turn, Multi-turn, Parsing, Transformation) & 34.82\% \\
    Mathematical Reasoning & 8.22\% \\
    General Tasks (Combination) & 33.36\% \\
    \bottomrule
    \end{tabular}
    }
    \caption{
    Data distribution of the Supervised Fine-tuning (SFT) process.
  }
  \label{tab:sft_data}
\end{table*}

This composition reflects a more streamlined view of the key categories of tasks, where coding-related tasks (Python and SQL) and table-related tasks (including single-turn and multi-turn interactions, data parsing, and transformation) have been consolidated for clarity. This ensures the SFT data covers diverse database- and BI-related tasks while focusing on realistic applications involving coding, table operations, and complex reasoning. The largest portion remains in general tasks, which integrates multiple task types to enhance \ours{}’s versatility in handling various analysis workflows.

\subsection{Data Cleaning}
The data cleaning phase is crucial and continues throughout the training process, ensuring the quality and relevance of the data used. The goal is to remove irrelevant, inconsistent, or erroneous data while preserving the core information that enhances the performance of the model for the wide application scenarios. Here’s a breakdown of the most critical aspects.

\subsubsection{Table Selection and Column Manipulation}
To ensure high-quality and consistent data for the BI application scenarios, we apply a systematic set of rules for table filtering and selection. Below are the key filtering rules we follow:

\begin{itemize}
    \item \textbf{remove horizontal tables}: Any tables structured horizontally are transposed or removed to maintain consistency in data orientation.
    \item \textbf{remove duplicate columns or columns ignoring case sensitivity}: Columns with identical names or names that differ only by case (e.g., ``Column'' and ``column'') are merged or discarded.
    \item \textbf{remove columns containing only underscores}: Columns that only contain underscores or similar meaningless data are removed.
    \item \textbf{remove columns where field content exceeds 100 characters}: Columns containing overly long text fields, which may be irrelevant or quite uncommon realistically, are filtered out.
    \item \textbf{remove rows or columns with more than 30\% \texttt{NaN} Values}: Any tables where a significant portion of the data is missing are filtered out to ensure data completeness.
    \item \textbf{remove tables with fewer than 5 rows or 2 columns}: Tables with insufficient data (less than 5 rows or 2 columns) are considered too sparse and are discarded.
    \item \textbf{remove columns where the first value directly matches the column name}: When a column’s first entry repeats its own column name, it’s a sign of an error or redundant information, so the column is removed.
    \item \textbf{more to be omitted in this report...}
\end{itemize}

These rules ensure that tables used in our model training are relevant, clean, and structured, optimizing the quality of the data pipeline for better performance.

\subsubsection{Table-query-answer Tuple Cleaning}

The tuple cleaning process follows a systematic pipeline to ensure the quality and accuracy of the table-query-answer data used in our model training. The pipeline is structured as follows:

\begin{center}
    \textbf{Pipeline}: Rule-based Filtering $\rightarrow$ Score-based Filtering $\rightarrow$ Sampling Calibration $\rightarrow$ Evaluation $\rightarrow$ Manual Review
\end{center}

The detailed stages within this pipeline are described below:

\begin{enumerate}
    \item \textbf{rule-based filtering}
    \begin{enumerate}
        \item \textbf{executable code and execution accuracy}: We developed \texttt{python-executor} and \texttt{sql-executor} scripts to ensure the executable nature of the code combined with table data. During this process, we filter out common errors such as field mismatch (key-error) or data type conversion errors that occur during execution.
        \item \textbf{regular expression-based filtering}: Using regular expressions and heuristic rules, we remove data exhibiting typical anomalies, including but not limited to:
        \begin{itemize}
            \item Instances where the generated content includes redefined custom table data that does not match the input.
            \item Outputs where the expected code type (e.g., Python) does not match the requested type (e.g., SQL), or vice versa.
            \item Outputs that contain only comment lines without any executable code.
        \end{itemize}
    \end{enumerate}

    \item \textbf{score-based filtering}
    \begin{enumerate}
        \item \textbf{score filtering method}: We apply multiple prompts and utilize several models (e.g., \texttt{gpt4o}, \texttt{gpt4-o1}) in different combinations to score the input-output pairs. The prompts target different aspects of the data such as input accuracy, code detail, and output clarity, to avoid overly complex prompt structures that may result in inaccurate scoring. Only samples scoring above a single threshold are retained.
        \item \textbf{score filtering pipeline}: The filtering process involves selecting samples based on \texttt{queries + table-info} as input, followed by filtering using the complete \texttt{queries + table-info + code/text-output} structure.
    \end{enumerate}

    \item \textbf{sampling calibration:}
        For executable code, we manually sample between 200-500 tuples. These samples are manually verified for both code accuracy and correct execution. If the sampling accuracy falls below a threshold (0.95), the responsible team is required to revise and optimize the data generation and score-filtering scripts, ensuring the quality of batch-generated data and its screening process.

    \item \textbf{evaluation:}
        We use a fixed validation set to evaluate the executable nature of LLM-generated outputs.

    \item \textbf{manual review and summary:}
        Human reviewers manually inspect the validation set for bad cases, checking for consistent errors. This step helps identify potential dirty data in the training batch and spot missing or needed data types (e.g., function-call capabilities, multi-turn response accuracy, or \texttt{React}-style task calibration).

\end{enumerate}

\subsection{Data Augmentation}

Data augmentation plays a critical role in improving the robustness of our training datasets. We employ several strategies to diversify the input and enhance the model's ability to handle varied data scenarios. The key augmentation methods are as follows:

\begin{itemize}
    \item \textbf{diverse table serialization input formats}: We utilize multiple formats for serializing table input data, including \texttt{html}, \texttt{xml}, \texttt{df.markdown()}, \texttt{df.to\_string()}, \texttt{json}, and various custom table serialization methods. These are randomly combined with table field descriptions, schema information, and value enumeration data, resulting in more than 20 different ways of constructing the input \texttt{table-info}.
    
    \item \textbf{query augmentation}: In BI-related queries, a significant number of queries tend to be ambiguous. To ensure that our model handles query ambiguity effectively, we aim to avoid direct mapping between fields in the query and table names or field names. Specifically, we ensure that quoted field or table names are removed from queries. Experimental results show that this method significantly reduces instances of key-errors in \ours{}, especially in realistic BI cases.
    
    \item \textbf{table data augmentation}: We anonymize table fields and categorical data values, enhancing the \ours{}’s robustness when handling dirty or noisy table data.
    
    \item \textbf{single-turn and multi-turn QA pair combination augmentation}: We combine single-turn and multi-turn QA data using different prompt formats and output structures. This reduces \ours{}'s sensitivity to specific system or user instructions, ensuring better generalization.
    
    \item \textbf{post-processing data augmentation}: During the data generation process, we randomly add details such as chart plotting instructions or summary explanations to create diverse outputs and further augment the dataset.
\end{itemize}

\subsection{SFT with the Encoder}
\label{sec:sft_enc}

Our supervised fine-tuning (SFT) process is generally divided into two primary scenarios. The first scenario involves standard SFT with input-output data, which covers most general-purpose tasks that do not involve structured tables. These tasks typically involve common textual inputs and outputs, where the focus is on aligning user queries with generated responses. The details of these general types of datasets can be found in Table~\ref{tab:sft_data}. 

The second scenario deals with table-related data, where we combine the decoder and encoder using a query-based transformer mechanism inspired by Q-Former~\cite{li2023blip}. This mechanism is specifically trained on table-query-answer tuples, which allows the model to process structured data efficiently. In this setup, the encoder processes table information, generating embeddings that are then combined with the query, aiding the decoder in producing accurate results.

Noticed that the encoder are first pretrained in a self-supervised manner prior to the SFT process, focusing on tasks such as schema inference, which helps the model understand general table semantics more effectively. 
While these two SFT scenarios (general and table-related) are distinct in their approach, they are combined during the training process. This combined approach allows the model to handle both free-form text queries and structured table data, making it flexible and adaptable across various BI tasks while keeping the general abilities in \ours{}. By dynamically switching between these two modes, the model can effectively process both unstructured and structured information depending on the task at hand.

Finally, we experimented with several training configurations to optimize the SFT process. In some cases, we froze the decoder and only trained the encoder and projection layers. In other setups, we trained the entire model end-to-end, which allowed us to learn simultaneously from both general and table-related data. These different training configurations provided valuable insights into how to best balance the model's training between its encoder and decoder components.
\section{Agent Framework}
\label{sec:agent}
To seamlessly integrate the \ours{} model with enterprise-level data analysis tools, we designed the agent workflow runtime framework. This framework consists of three core components: runtime prompt engineering, code sandbox, and an independent agent evaluation module. Together, these elements enhance the agent’s intelligent data analysis capabilities and reliability, while ensuring that the generated code is executed safely and evaluated automatically. This structure allows for efficient management and monitoring of the agent’s overall performance.

\begin{figure*}[!t]
    \centering
    \includegraphics[width=1.0\linewidth]{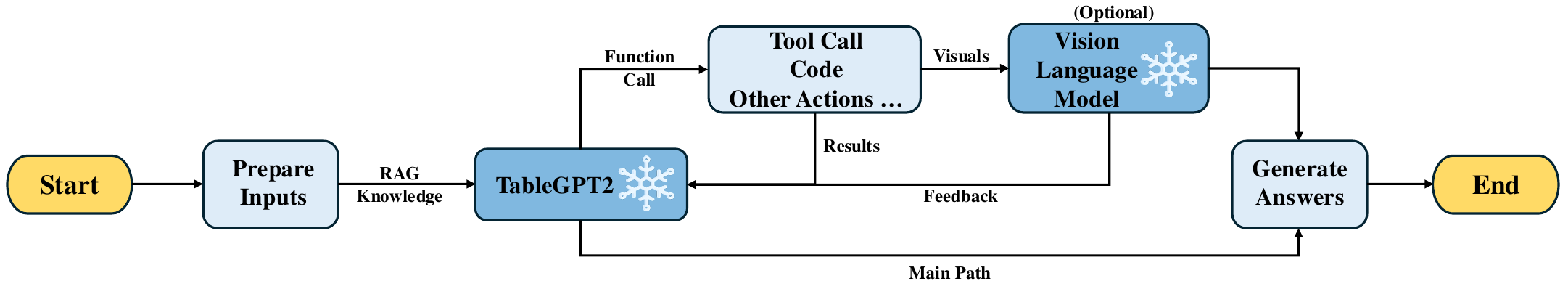}
    \caption{
    The workflow of the complete agent process. Input queries are initially prepared and processed through a prompt engineering module. With the assistance of a retrieval-augmented generation (RAG) module from an external knowledge base, they are fed into the main model. \ours{} then collaborates with a visual language model (VLM) to generate tool calls, code, or other related actions. By observing intermediate results, the execution may iterate as necessary, leveraging the agent's reflection ability. This iterative process ultimately leads to the final output $y$ through seamless interaction between agents and tools.
    }
    \label{fig:agent_workflow}
\end{figure*}

The agent workflow runtime aims to automate support for complex data analysis tasks by integrating data warehouse connection, visualization, and analysis functions into a cohesive workflow. The core execution process, as shown in the Figure~\ref{fig:agent_workflow}, is divided into the following key steps:

\begin{itemize}
    \item 
        \textbf{input table normalization (optional):} This step ensures that the input dataset is in a format easily processed by \ours{}. Although \ours{} is designed to handle various types of tabular data, certain formats—such as those with merged cells or multiple headers—can present challenges for LLM interpretation. In such cases, the dataset undergoes normalization, which involves two key processes:
        \begin{itemize}
            \item \textbf{table transformation:} This process identifies issues such as merged cells or multiple headers in the input dataset. When such features are present, the dataset is adjusted accordingly, for example, by merging headers or unifying merged cells into a standardized structure.
            \item \textbf{transformation-related code generation:} Both the original and transformed tables are used to generate code that automates the transformation process. This code can then be applied to normalize subsequent states in the same conversation thread.
    \end{itemize}
    \item \textbf{prepare inputs:} this step parses and processes the content retrieved via RAG, supportive examples and other supplementary information, providing the agent with the necessary context.
    \item \textbf{agent execution:} This step is the core component of the workflow. The input is first processed by the large language model agent, which generates code based on the given instructions. Depending on the task, two agents with different functionalities are employed:
    \begin{itemize}
        \item \textbf{large language model:} The LLM is the primary agent responsible for generating code. Based on the input instructions, the agent performs tasks such as:
            \begin{itemize}
                \item \textbf{SQL/Python code generation:} When the input involves data retrieval (e.g., from a warehouse) or data visualization tasks, the agent generates SQL queries to fetch the data and Python code to process or visualize it (e.g., creating charts or graphs). This unified approach ensures that data retrieval and visualization are seamlessly integrated.
            \end{itemize}
        \item \textbf{vision language model (optional):} The VLM is an optional agent activated when figures are part of the task, offering two key actions:
            \begin{itemize}
                \item \textbf{chart analysis:} For figures generated by visualization tools, the agent analyzes charts and visual outputs to provide insights into data features, trends, and anomalies.
                \item \textbf{feature learning:} This action allows the agent to learn from the generated figures, extracting complex features that extend beyond basic code generation, thereby enhancing the workflow's capacity to tackle more intricate tasks.
            \end{itemize}
    \end{itemize}
    \item \textbf{tool call:} the agent invokes specific tools (e.g., database connections, file operations, and data visualization, etc.) and continues the workflow based on the returned results.
\end{itemize}

Through this workflow design, the agent workflow runtime effectively supports the entire process of data analysis, visualization, and interpretation, providing a clear analysis path even in complex data environments.

\subsection{Prompt Engineering}

The prompt engineering primarily involves the design of effective system instructions and the integration of RAG infrastructure.

\subsubsection{System Instruction}

The system instruction is used to define the agent's role in the dialogue, the scope of the analysis, and how to organize the format of the answers. In designing the System Instruction, we adopted a layered instruction strategy, gradually incorporating analysis goals, response style, output format, and context management into the instructions to improve the agent's response consistency and accuracy. The main goals of the instructions are as follows:

\begin{itemize}
    \item \textbf{role definition:} by introducing a clear role definition (as a data analysis expert), we ensure that the agent maintains professionalism in all dialogues.
    \item \textbf{functional guidance:} the instructions specify that the agent should analyze data and generate Python code to perform the tasks. The generated code is executed in an iPython environment, and based on the results, the agent provides further analytical suggestions.
    \item \textbf{usage guidelines:} this includes specific functions for reading data, preferences for charting tools (e.g., Seaborn), and strategies for handling errors encountered during execution.
    \item \textbf{limitations:} for non-data-related questions (e.g., politically sensitive or privacy-related issues), the agent is instructed to refuse to answer, ensuring controlled behavior in complex scenarios.
\end{itemize}

The design of this system instruction aims to guide the agent in producing responses that align with real-world applications, meeting business requirements while ensuring safety and compliance.

\subsection{Retrieval-Augmented Generation}

Enterprise-level data warehouses often contain petabytes (PB) of data and complex schemas with thousands of tables. Even though traditional large language models (LLMs) possess large context-handling capabilities, they are still unable to process such vast amounts of data at once. Inputting excessive context into the model can lead to distracted attention, significantly affecting both the accuracy and the response speed of the model. Therefore, RAG technology~\cite{lewis2020retrieval} plays a crucial role in such scenarios.

When dealing with large datasets and complex database schemas, our RAG engine first retrieves enterprise-specific data (such as database schemas and business documents) and selects highly relevant contextual information for the current analysis task through two stages: embedding and reranking. This avoids unnecessary information overload. The agent then generates code or interprets the data based on this refined context, focusing on the most important data and analysis goals within the limited context window.

In data analysis tasks, RAG is applied in two primary scenarios:

\begin{itemize}
    \item \textbf{data schema retrieval and context aggregation:} when facing multiple tables or complex database structures, the agent uses RAG to retrieve table structures, field information, and relationship dependencies that are relevant to the current analysis task, and then generates the appropriate code. This approach effectively reduces unnecessary context loading, allowing the agent to quickly pinpoint data sources and generate more accurate code.
    
    \item \textbf{domain-specific knowledge retrieval and analysis:} in scenarios where the analysis involves specific business contexts (e.g., interpreting enterprise documents or historical analysis reports), RAG retrieves relevant document fragments and merges them with the current task context. This enables the agent to leverage domain-specific knowledge, improving the professionalism and credibility of its responses.
\end{itemize}

By integrating RAG technology, the agent can intelligently understand context when faced with complex data warehouses or multi-dimensional business needs. This allows it to generate analysis code or interpret data in a way that aligns with business requirements, significantly improving its usefulness and reliability in enterprise-level data environments.

\subsection{Code Sandbox}

In the agent workflow, the code generated by the agent needs to be executed in a safe, isolated, and controlled environment. To ensure that the execution of the code does not pose a security threat to the main system and can be quickly isolated and recovered in case of an error, we designed and implemented a specialized \textbf{code sandbox} environment. The code sandbox has the following core design objectives:

\begin{itemize}
    \item \textbf{security:} prevent unauthorized operations on the host system or other critical services by the generated code.
    \item \textbf{isolation:} ensure that each code execution environment is independent and does not share any system resources or files.
    \item \textbf{easy recovery:} in case of errors or improper behavior, the environment can be quickly destroyed and recreated without affecting other running tasks.
\end{itemize}

We adopt a containerized approach to package the iPython executor as a code sandbox, using \textbf{Kubernetes} as the management platform for these sandboxes. Each piece of code generated by the agent is executed within an independent sandbox, providing the following safeguards:

\begin{itemize}
    \item \textbf{isolated execution environment:} each container has its own isolated file system, process space, and network space, ensuring no direct impact on the host system.
    \item \textbf{resource quota control:} Kubernetes' resource quota functionality (such as CPU and memory limits) is used to prevent any single task from consuming excessive resources.
    \item \textbf{timeout execution strategy:} all code execution tasks are set with a maximum execution time (timeout), after which the environment is automatically terminated and destroyed.
\end{itemize}

The data analysis tasks typically consist of multiple steps, with potential dependencies between each step. Therefore, the lifecycle of a code sandbox must cover the entire duration of a complete data analysis task. However, enterprise-level data analysis tools often handle large numbers of independent analysis tasks, each executed in its own sandbox, which can consume significant cluster resources. To ensure efficient utilization of resources, we designed an \textbf{automatic destruction mechanism} for the code sandbox. When a sandbox has been idle for an extended period, the container is automatically destroyed to free up resources.

We have made the entire agent workflow open-source on our Github repository. See Section~\ref{sec:disclaimer}.
\section{Evaluation of \ours{}}
\label{sec:exp}
To effectively evaluate \ours{}, we constructed a comprehensive benchmark for table-related tasks, integrating a diverse range of existing datasets along with a newly collected, realistic, and complex table dataset. Spanning 6 major tasks across 7 domains, this benchmark offers a rigorous assessment platform. 

For our comparisons, we selected a range of large language models (LLMs), both those tailored and those not specifically tuned for table- or BI-related tasks. The first category consists of the most advanced open-source general-purpose LLMs, including \textbf{DeepSeek-Coder-V2-Lite-16B}~\cite{liu2024deepseek}, \textbf{Yi-Coder-9B-Chat}\cite{yicoder}, \textbf{Qwen2.5-Coder-7B-Instruct}~\cite{hui2024qwen25codertechnicalreport}, and \textbf{Qwen2.5-7B-Instruct}~\cite{qwenteam2024qwen25}.
The second category comprises models fine-tuned or explicitly developed for table-related tasks. This includes \textbf{TableLLMs}~\cite{zhang2024tablellm, wu2024tablebench}, a set of large language models specifically designed and optimized for table analysis. In particular, \cite{zhang2024tablellm} fine-tuned \textbf{CodeLlama-13B}~\cite{roziere2023code} to handle a variety of table operations in spreadsheet and document settings, with an emphasis on real-world tabular data manipulation. Additionally, \cite{wu2024tablebench} constructed a comprehensive TableQA instruction corpus to fine-tune robust baseline models across various LLMs—including \textbf{Qwen2-7B}~\cite{yang2024qwen2}, \textbf{CodeQwen-7B}~\cite{qwen}, \textbf{Llama3-8B}~\cite{dubey2024llama}, \textbf{Llama3.1-8B}~\cite{dubey2024llama}, and \textbf{Deepseek-Coder-7B}~\cite{guo2024deepseek}—in order to assess reasoning capabilities within tabular data contexts.

\paragraph{Three important notes.} 
We further note that: 
\begin{enumerate}
    \item To provide clarity, our experimental comparisons are organized chronologically by the release timeline of these models, as LLM performance generally improves with each new generation.
    \item As noted in recent research~\cite{anonymous2024rethinking}, some LLMs fine-tuned specifically for tabular data, such as \textbf{TableLlama}~\cite{zhang2024tablellama}, exhibit significant drops in performance on general benchmarks. We have excluded these models from our evaluation due to their overfitting on table-related tasks at the expense of broader generalization capabilities.
    \item \textbf{Benchmark neutrality}: the objective of \ours{}, like many other general-purpose LLMs, is not to overfit any single benchmark. We exclude specialized methods designed to excel on specific benchmarks, such as CHASE-SQL~\cite{pourreza2024chasesqlmultipathreasoningpreference} on MCS-SQL~\cite{lee2024mcs}, due to their narrow focus. \ours{} does not undergo repeated training on samples from these benchmarks, ensuring a balanced and versatile model performance.
\end{enumerate}

\subsection{Benchmark Overview}

\subsubsection{Data Setups}

\paragraph{Existing benchmarks.}

\begin{table*}[]
\small
\centering
\resizebox{\linewidth}{!}{
\begin{tabular}{@{}llllcccc@{}}
\toprule
\multirow{2}{*}{Task Category} &\multirow{2}{*}{Task Description} & \multirow{2}{*}{Benchmark} &\multirow{2}{*}{Metric}  & \#Test & \multicolumn{3}{c}{Input Token Length}\\ 
& & & & (Table/Sample) & min & max & median \\
\midrule
\multirow{4}{*}{  
\begin{tabular}[c]{@{}l@{}}Table\\ Understanding\end{tabular}
} & Interpretation & Column Type Annotation & F1 & 1K/2K & 1650 & 7065 & 2152\\ 
& Interpretation &Relation Extraction & F1 & 1K/2K & 1910 & 6914 & 2670\\ 
& Interpretation &Entity Linking & Acc & 1K/2K & 385 & 7736 & 2114\\ 
& Augmentation & Row Population & MAP & 0.3K/0.3K & 686 & 6712 & 2059\\ \midrule
\multirow{5}{*}{ 
\begin{tabular}[c]{@{}l@{}}Question\\ Answering\end{tabular}
} & Hierarchical Table QA &HiTab & Exec Acc & 1K/1K & 229 &5501 & 847\\
& Free-form Table QA &FeTaQA & BLEU & 2K/2K & 236 & 3049 & 591\\ 
&Hybrid Table QA & HybridQA & Acc & 3K/3K & 276 & 2365 &675\\
&Table QA &WikiSQL & Acc & 5K/16K & 204 & 2034 & 566\\
&Table QA &WikiTQ & Acc & 0.4K/4K & 266 & 2120 & 691\\
\midrule
\multirow{2}{*}{ 
\begin{tabular}[c]{@{}l@{}}Fact\\ Verification\end{tabular}
} & \multirow{2}{*}{Fact Verification}& TabFact &Acc & 2K/12K & 246 & 2861 & 578\\ 
&& FEVEROUS &Acc & 4K/7K & 262 & 8192 & 654\\
\midrule
\multirow{1}{*}{ 
\begin{tabular}[c]{@{}l@{}}Table to Text\end{tabular}
} &  
\begin{tabular}[c]{@{}l@{}}Highlighted\\ Cells Description \end{tabular}
& ToTTo &BLEU & 7K/8K & 139 & 8192 & 210\\ \midrule
\multirow{4}{*}{ 
\begin{tabular}[c]{@{}l@{}}Natural Language\\ to SQL\end{tabular}
} & Hierarchical Table QA &BIRD (dev) & Exec Acc & 89/1.5K & 290 &2131 & 962\\
& Highlighted Cells QA &BIRD (knowledge) & Exec Acc & 89/1.5K & 290 & 2248 & 985\\ 
&Hybrid Table QA & Spider (dev) & Exec Acc & 83/1K & 171 & 950 & 321\\
&Table QA &Spider (test) & Exec Acc & 188/2.1K & 134 & 1083 & 297\\ \midrule
\multirow{1}{*}{ 
\begin{tabular}[c]{@{}l@{}}Holistic Evaluation\end{tabular}
} &  
\begin{tabular}[c]{@{}l@{}}Highlighted\\ Cells Description \end{tabular}
& TableBench & \begin{tabular}[c]{@{}l@{}}Rouge-L /\\Pass@1 \end{tabular}
& 0.59K/3.5K & 324 & 8192 & 776\\
\bottomrule
\end{tabular}
}
\caption{Statistics of tasks and datasets on the existing benchmark.}
\label{tab:benchmark}
\end{table*} 

As shown in Table~\ref{tab:benchmark}, we compiled a comprehensive collection of existing benchmarks for table understanding and reasoning, encompassing 27.7K tables and 88.9K test samples. The specific data volumes and distribution information (e.g. input token lengths across test samples) are detailed in Table~\ref{tab:benchmark}.
Following prior works, we further divided the benchmarks into 6 primary table analysis tasks, enabling a comprehensive evaluation of our model's performance across varied task types:

\begin{itemize}
    \item \textbf{table understanding:} TURL~\cite{deng2020turl} proposed a table understanding benchmark consisting of 4 widely studied tasks covering table interpretation (e.g., column type annotation, relation extraction, entity linking) and table augmentation (e.g., row population). We employed Micro-F1, accuracy, and mean average precision (MAP) as evaluation metrics. These benchmarks evaluate the model's basic understanding ability on tables. 
    \item \textbf{table question answering (TableQA):} TableQA focuses on answering natural language questions based on tabular data. We have collected standard flat relational table QA datasets, such as WikiTableQuestion (WikiTQ)~\cite{wikitq} and WikiSQL~\cite{wikisql}, as well as complex hierarchical table QA datasets like HiTab~\cite{cheng-etal-2022-hitab}. Additionally, we include free-form table QA from FeTaQA~\cite{Nan2021FeTaQAFT} and multi-hop reasoning involving hybrid table and text data from HybridQA~\cite{chen2020hybridqa}. We utilized Execution accuracy~\cite{wikisql}, BLEU, and accuracy as the key evaluation metrics. These benchmarks test the model's ability to interpret and retrieve information from complex tables to provide accurate answers.
    \item \textbf{table fact verification:} We included datasets like TabFact~\cite{Chen2020TabFact} and FEVEROUS~\cite{feverous}, which assess the model's ability to verify the truthfulness of statements against the information presented in tables. In all cases, accuracy was used as the evaluation metric. This requires logical reasoning and fact-checking skills to confirm or refute claims based on tabular data.
    \item \textbf{table-to-text generation (Table2Text):} We utilized datasets such as ToTTo~\cite{parikh-etal-2020-totto}, which aim at generating coherent and informative textual descriptions from structured table data. BLEU was employed as the evaluation metric. This task measures the model's ability to summarize and verbalize tabular information effectively.
    \item \textbf{natural language to SQL (NL2SQL):} Benchmarks like BIRD~\cite{li2024bird} and Spider~\cite{yu2018spider} were incorporated to evaluate the model's capability to translate natural language queries into executable SQL statements for database querying. \texttt{BIRD (knowledge)} denotes the reasoning mechanism for external knowledge grounding between natural language questions and database values provided by the dataset~\cite{li2024bird}, facilitating the assessment of LLMs in utilizing external knowledge bases for reasoning. Execution accuracy was consistently used as the evaluation metric. This task assesses the model's understanding of linguistic nuances and its proficiency in generating correct SQL code.
    \item \textbf{holistic evaluation:} 
    To comprehensively evaluate \ours{}, TableBench~\cite{wu2024tablebench} is incorporated, a manually annotated TableQA benchmark with 886 samples across 18 industrial domains, supporting four key table analysis tasks: fact verification, numerical reasoning, data analysis, and code-based chart visualization. Various reasoning methods such as symbolic chain-of-thought (SCoT), textual chain-of-thought (TCoT), program-of-thought (PoT), and direct prompting (DP) strengthen LLM reasoning abilities, creating nearly 3.5k test instances. Only PoT and chat visualization require code generation evaluated by pass@1~\cite{chen2021evaluating}, while other methods use Rouge-L. This dataset assesses complex reasoning capabilities of LLMs in real-world scenarios.
\end{itemize}

\paragraph{A new benchmark: RealTabBench.}
Despite the abundance of existing benchmarks, most focus on relatively simple tasks and fail to align with real-world usage scenarios. For example, in TableQA tasks, the questions often involve straightforward data retrieval, and benchmarks for irregular tables typically assess only the hierarchical structure of table headers. To address these limitations, we have constructed a new benchmark that is both more challenging and more representative of practical applications. Building upon the aforementioned tasks, we collected 360 complex data tables from real-world tables in BI scenarios and formulated 6,000 realistic and intricate query sentences based on them.

Notably, we have specifically examined two table characteristics that are particularly troublesome in practical applications:
\begin{itemize}
    \item \textbf{ambiguity}: The inherent ambiguity in table content poses significant challenges for automated analysis. Our benchmark includes tables with ambiguous entries to assess the model's ability to interpret and resolve such ambiguities effectively.

    \item \textbf{irregularity}: In production settings, tabular data often includes pervasive merge operations and irregular structures, such as merged cells and non-uniform layouts. While existing benchmarks, such as HiTab, typically focus on hierarchical table headers, our enhanced benchmark incorporates tables with irregular layouts and complex formatting to evaluate the model's capability to handle non-standard table formats commonly encountered in real-world scenarios.
\end{itemize}

On RealTabBench, the generated results are evaluated across three key dimensions:

\begin{itemize}
\item \textbf{consistency}: Measures whether the model’s answer is semantically aligned with the reference answer, ensuring accurate and unbiased understanding of the question.
\item \textbf{information completeness}: Assesses whether the model’s answer includes all essential information and analytical steps, avoiding any critical omissions.
\item \textbf{security}: Evaluates potential security risks in the response, ensuring the target model does not approach or exceed sensitive boundaries.
\end{itemize}

A hybrid system, combining human reviewers and an Evaluation LLM, is employed to produce the final scores. 
We have made our entire evaluation pipeline with a portion of samples publicly available\footnote{\url{https://github.com/tablegpt/tablegpt-agent/tree/main/realtabbench}}.

\begin{table*}[!t]
  \centering
  \resizebox{\linewidth}{!}{%
  \begin{tabular}{ll|c|cccccccccccc}
    \toprule
    \textbf{Benchmark} & \textbf{Metric}  & GPT-4o & \begin{tabular}[c]{@{}c@{}}TableLLM\\ (Qwen2)\end{tabular} & \begin{tabular}[c]{@{}c@{}}TableLLM\\ (CodeQwen)\end{tabular} & \begin{tabular}[c]{@{}c@{}}TableLLM\\ (LLaMA3)\end{tabular} & \begin{tabular}[c]{@{}c@{}}TableLLM\\ (LLaMA3.1)\end{tabular} &\begin{tabular}[c]{@{}c@{}}TableLLM\\ (DeepSeek)\end{tabular} & TableLLM-13B & DeepSeek-lite & Yi-Coder & Qwen2.5-Coder & Qwen2.5-Instruct & \textbf{\ours{}-7B} & \textbf{\ours{}-72B} \\
    \midrule
    \multicolumn{14}{c}{\textit{Table Understanding}} \\
    \midrule
    Col Type Annot. & F1 & 31.75 & 10.10 & 5.71 & 1.47 & 1.59 & 6.04 & 12.70 & 20.58 & 5.38 & 32.59 & 22.19 & \cellcolor{mygray}{\textbf{85.88}} & \cellcolor{mygray}{85.67} \\
    Relation Extract. & F1 & 52.95 & 1.60 & 3.79 & 2.39 & 2.00 & 3.34 & 18.16 & 8.67 & 2.25 & 31.00 & 15.92 & \cellcolor{mygray}{\textbf{83.35}} & \cellcolor{mygray}{79.50} \\
    Entity Linking & Acc & 90.80 & 47.10 & 39.70 & 0.20 & 0.60 & 15.50 & 66.25 & 70.15 & 41.75 & 71.70 & 82.25 & \cellcolor{mygray}{92.00} & \cellcolor{mygray}{\textbf{93.30}} \\
    Row Pop. & MAP & 53.40 & 2.20 & 5.14 & 1.93 & 6.23 & 3.13 & 14.25 & 1.20 & 1.00 & 13.23 & 12.30 & \cellcolor{mygray}{\textbf{59.97}} & \cellcolor{mygray}{55.83} \\
    \midrule
    \multicolumn{14}{c}{\textit{Question Answering}} \\
    \midrule
    HiTab & Exec Acc & 48.40 & 11.74 & 0.00 & 0.00 & 0.00 & 39.08 & 6.30 & 0.76 & 0.00 & 1.70 & 10.73 & \cellcolor{mygray}{70.27} & \cellcolor{mygray}{\textbf{75.57}} \\
    FetaQA & BLEU & 21.70 & 12.24 & 8.69 & 2.42 & 3.10 & 7.94 & 10.83 & 15.08 & 11.17 & 13.00 & 16.91 & \cellcolor{mygray}{28.97} & \cellcolor{mygray}{\textbf{32.25}} \\
    HybridQA & Acc & 58.60 &  27.12 & 20.14 & 27.35 & 27.61 & 19.53 & 51.88 & 42.58 & 29.83 & 51.10 & 51.13 & \cellcolor{mygray}{53.17} & \cellcolor{mygray}{\textbf{56.41}} \\
    WikiSQL & Acc & 47.60 & 46.50 & 37.20 & 39.26 & 39.00 & 36.14 & 41.10 & 38.30 & 25.34 & 46.90 & 47.42 & \cellcolor{mygray}{53.74} & \cellcolor{mygray}{\textbf{57.32}} \\
    WikiTQ & Acc & 68.40 & 64.16 & 36.05 & 34.95 & 38.84 & 36.05 & 66.30 & 47.65 & 43.37 & \textbf{74.50} & 68.55 & \cellcolor{mygray}{61.42} & \cellcolor{mygray}{71.45} \\
    \midrule
    \multicolumn{14}{c}{\textit{Fact Verification}} \\
    \midrule
    TabFact & Acc & 74.40 & 72.00 & 53.20 & 40.06 & 27.13 & 60.76 & 68.95 & 62.27 & 79.6 & 77.26 & 84.60 & \cellcolor{mygray}{77.80} & \cellcolor{mygray}{\textbf{85.43}} \\
    FEVEROUS & Acc & 71.60 &  20.10 & 46.90 & 51.50 & 42.30 & 18.39 & 21.45 & 7.80 & 38.10 & 60.70 & 63.30 & \cellcolor{mygray}{\textbf{78.05}} & \cellcolor{mygray}{76.80} \\
    \midrule
    \multicolumn{14}{c}{\textit{Table to Text}} \\
    \midrule
    ToTTo & BLEU & 12.21 & 6.95 & 3.10 & 5.50 & 6.23 & 3.81 & 5.36 & 8.76 & 2.64 & 10.50 & 11.91 & \cellcolor{mygray}{14.10} & \cellcolor{mygray}{\textbf{22.69}} \\
    \midrule
    \multicolumn{14}{c}{\textit{Natural Language to SQL}} \\
    \midrule
    BIRD(dev) & Exec Acc & - & 9.13 & 7.37 & 1.83 & 2.48 & 0.39 & 0.72 & 25.10 & 24.19 & 27.18 & 18.97 & \cellcolor{mygray}{31.42} & \cellcolor{mygray}{\textbf{38.40}} \\
    BIRD(dev-knowledge) & Exec Acc & - & 15.45 & 18.19 & 3.39 & 3.72 & 0.39 & 1.83 & 36.51 & 39.96 & 42.96 & 31.42 & \cellcolor{mygray}{49.28} & \cellcolor{mygray}{\textbf{60.76}} \\
    Spider(dev) & Exec Acc & - & 42.26 & 32.88 & 12.86 & 18.96 & 2.71 & 4.26 & 66.44 & 58.12 & 70.99 & 61.70 & \cellcolor{mygray}{76.31} & \cellcolor{mygray}{\textbf{79.40}} \\
    Spider(test) & Exec Acc & - & 40.29 & 34.93 & 12.02 & 16.35 & 7.33 & 2.93 & 66.65 & 56.87 & 69.73 & 60.18 & \cellcolor{mygray}{74.38} & \cellcolor{mygray}{\textbf{78.48}} \\
    \midrule
    \multicolumn{14}{c}{\textit{Holistic Table Evaluation}} \\
    \midrule
    \multirow{4}{*}{TableBench} & DP & - & 26.62 & 26.44 & 26.71 & 26.73 & 26.15 & 3.88 & 29.60 & 21.94 & 28.67 & 25.18 & \cellcolor{mygray}{32.03} & \cellcolor{mygray}{\textbf{38.90}} \\
    & TCoT & - &  37.08 & 31.33 & 29.79 & 30.01 & 28.65 & 3.85 & 30.93 & 22.8 & 36.25 & 29.77 & \cellcolor{mygray}{42.34} & \cellcolor{mygray}{\textbf{50.06}} \\
    & SCoT & - & 14.11 & 17.78 & 9.60 & 12.38 & 22.39 & 2.88 & 22.61 & 8.43 & 25.95 & 24.35 & \cellcolor{mygray}{25.01} & \cellcolor{mygray}{\textbf{30.47}} \\
    & PoT@1 & - & 21.05 & 26.39 & 31.96 & 25.80 & 28.39 & 2.94 & 10.90 & 11.36 & 16.15 & 22.58 & \cellcolor{mygray}{\textbf{33.52}} & \cellcolor{mygray}{28.98} \\
    \bottomrule
  \end{tabular}%
  }
  \caption{The experimental results on the existing table-related benchmarks. The bolded performances do not include GPT-4o.
  }
  \label{tab:main_results}
\end{table*}

\subsection{Results}
In Table~\ref{tab:main_results}, we present the primary comparisons of \ours{} (in both 7B and 72B versions) against the most advanced LLM competitors discussed above.

Notably, without extensive training on any specific benchmark training set, \ours{} achieves significantly better results than nearly all other LLM-based methods. Empirically, on certain benchmarks, \ours{} even reaches results superior or comparable to those of GPT-4o.

Moreover, on complex data benchmarks such as HiTab, which involve hierarchically structured tables, most current LLMs show poor performance. \ours{}, however, demonstrates a substantial improvement, achieving over a 60\% absolute increase in execution accuracy over Qwen2.5 family.

In addition, we present our results on the RealTabBench dataset, in Table~\ref{tab:realtabbench-results}.

\begin{table*}[htb]
  \centering
  \resizebox{\linewidth}{!}{%
  \begin{tabular}{l|cccccccccc|c}
    \toprule
    Task & \begin{tabular}[c]{@{}c@{}}TableLLM\\(Qwen2)\end{tabular} & \begin{tabular}[c]{@{}c@{}}TableLLM\\(CodeQwen)\end{tabular} & \begin{tabular}[c]{@{}c@{}}TableLLM\\ (LLaMA3)\end{tabular} & \begin{tabular}[c]{@{}c@{}}TableLLM\\ (LLaMA3.1)\end{tabular} &\begin{tabular}[c]{@{}c@{}}TableLLM\\ (DeepSeek)\end{tabular} & TableLLM-13B & DeepSeek-lite & Yi-Coder & Qwen2.5-Coder & Qwen2.5-Instruct & \textbf{\ours{}-7B} \\
    \midrule
    General & 44.72 & 40.84 & 33.92 & 30.05 & 20.82 & 22.52 & 41.05 & 51.58 & 69.27 & 74.99 & \cellcolor{mygray}{\textbf{75.79}} \\
    Irregular & 13.87 & 16.22 & 16.32 & 18.64 & 5.91 & 25.81 & 22.12 & 20.75 & 31.45 & 31.95 & \cellcolor{mygray}{\textbf{36.34}} \\
    Ambiguous & 38.20 & 37.91 & 28.81 & 27.01 & 21.25 & 21.95 & 37.84 & 47.68 & 61.32 & 65.78 & \cellcolor{mygray}{\textbf{68.33}} \\
    \bottomrule
  \end{tabular}%
  }
  \caption{
  The experimental results on our RealTabBench.
  }
  \label{tab:realtabbench-results}
\end{table*}

\begin{center}
\begin{figure}[!t]
  \makebox[\textwidth]{\includegraphics[width=0.9\paperwidth]{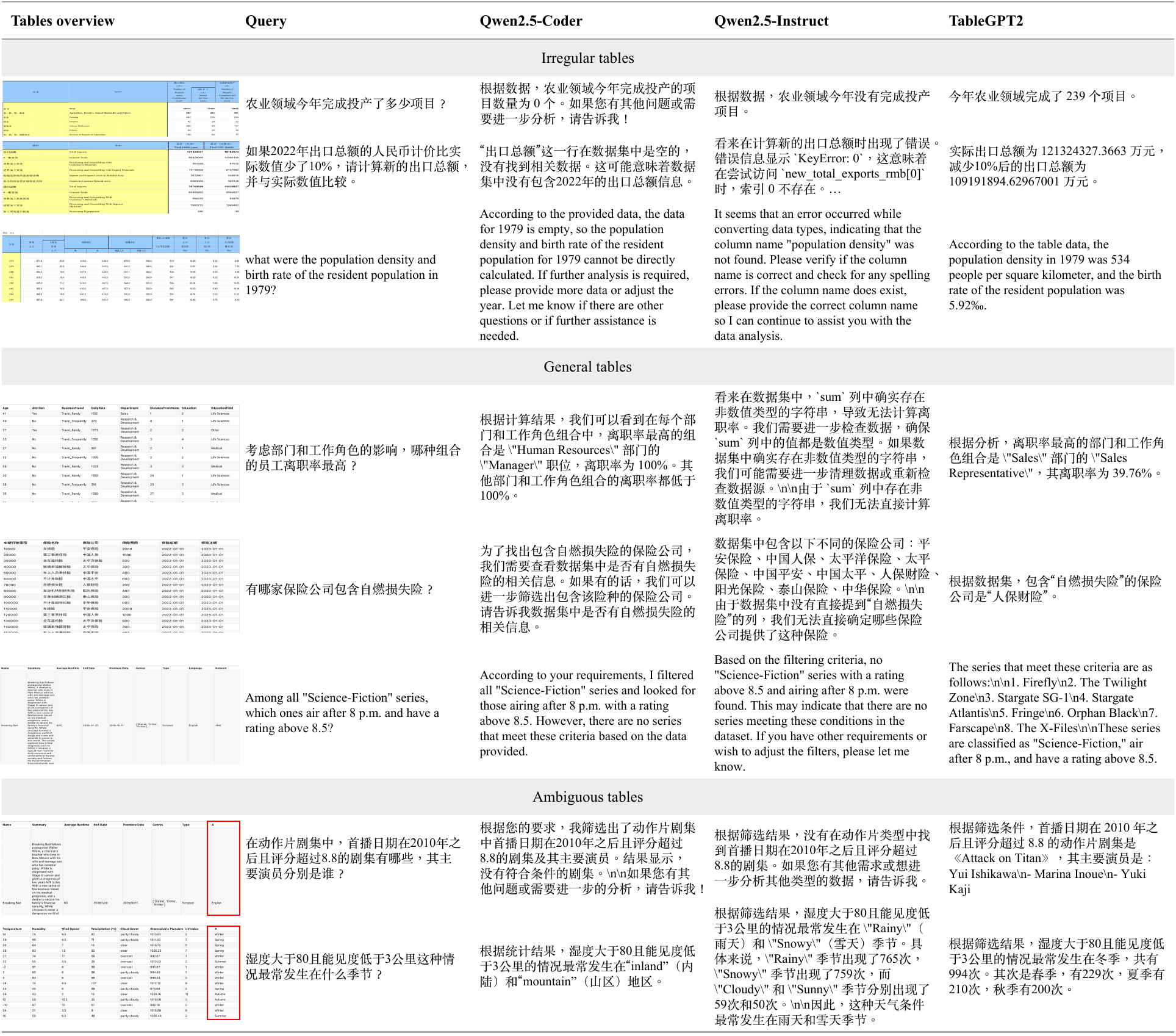}}
  \caption{Qualitative cases of \ours{}.}
  \label{fig:case_study}
\end{figure}
\end{center}

\subsection{Qualitative Results}
For qualitative illustration, we pick a few comparative cases in Figure~\ref{fig:case_study}, covering scenarios involving irregular tables, general tables, and ambiguous tables.

\subsection{Miscellaneous}
\begin{table*}[htb]
  \centering
  \resizebox{\linewidth}{!}{%
  \begin{tabular}{llcccccccccc|cc}
    \toprule
    \textbf{Benchmark} & \textbf{Metric} & \begin{tabular}[c]{@{}c@{}}TableLLM\\ (Qwen2)\end{tabular} & \begin{tabular}[c]{@{}c@{}}TableLLM\\ (CodeQwen)\end{tabular} & \begin{tabular}[c]{@{}c@{}}TableLLM\\ (LLaMA3)\end{tabular} & \begin{tabular}[c]{@{}c@{}}TableLLM\\ (LLaMA3.1)\end{tabular} &\begin{tabular}[c]{@{}c@{}}TableLLM\\ (DeepSeek)\end{tabular} & \multicolumn{1}{c|}{TableLLM-13B} & DeepSeek-lite & Yi-Coder & Qwen2.5-Coder & \multicolumn{1}{c|}{Qwen2.5-Instruct} & \textbf{\ours{}-7B} & \textbf{\ours{}-72B} \\
    \midrule
    \multicolumn{14}{c}{\textit{Code}} \\
    \midrule
    MBPP & Acc & 0.55 & 0.57 & 0.33 & 0.36 & 0.63 & 0.20 & 0.72 & 0.70 & 0.76 & 0.65 & \cellcolor{mygray}{0.64} & \cellcolor{mygray}{0.78} \\
    HumanEval & Acc & 0.62 & 0.58 & 0.21 & 0.15 & 0.63 & 1.83 & 0.78 & 0.85 & 0.86 & 0.84 & \cellcolor{mygray}{0.80} & \cellcolor{mygray}{0.84} \\
    \midrule
    \multicolumn{14}{c}{\textit{Language}} \\
    \midrule
    CMMLU & Acc & 79.75 & 26.51 & 25.55 & 25.18 & 48.45 & 7.70 & 60.98 & 44.76 & 66.71 & 80.07 & \cellcolor{mygray}{79.45} & \cellcolor{mygray}{88.91} \\
    MMLU & Acc & 65.97 & 31.03 & 26.45 & 26.29 & 49.49 & 1.80 & 55.68 & 50.95 & 64.31 & 73.71 & \cellcolor{mygray}{72.01} & \cellcolor{mygray}{84.27} \\
    \bottomrule
  \end{tabular}%
  }
  \vspace{5pt}
  \caption{
  The experimental results on benchmarks related to general ability.
  }
  \label{tab:general}
\end{table*}

As mentioned, tuning an LLM for table-related tasks should not compromise its overall capabilities. To verify this, we conducted evaluations on popular benchmarks, including MBPP~\cite{austin2021program}, HumanEval~\cite{chen2021evaluating}, CMMLU~\cite{li2023cmmlu}, and MMLU~\cite{hendrycks2020measuring}, which measure general coding and language comprehension skills (see Table~\ref{tab:general}). Unlike some other approaches highlighted in ~\cite{anonymous2024rethinking}, \ours{} maintains strong performance on these benchmarks, demonstrating no degradation in general competency.

\section{Outlook}

In this article, we introduced \ours{}, a model designed to address the integration of large language models (LLMs) into business intelligence (BI) workflows. However, despite achieving state-of-the-art (SOTA) performance in our experiments through careful design and implementation, \ours{} does not yet fully resolve the challenges of deploying LLMs in real-world BI environments. While significant progress has been made, there are still gaps that need to be addressed before it can be reliably used in production systems. In this section, we discuss some key techniques and approaches that could help bridge this gap.

\subsection{Coding in Specific Domains}
While we utilized Python and SQL data for fine-tuning \ours{}, specific domains often require specialized coding practices for security and efficiency reasons. For example, some industries employ pseudo-SQL or domain-specific languages (DSLs) designed to limit access or adhere to strict compliance standards. In such cases, it becomes essential to build interpreters dedicated to these specific domains, ensuring that the generated code integrates seamlessly into existing data infrastructures.

A key challenge here is enabling LLMs to quickly adapt to enterprise-specific DSLs or pseudo-code. Although LLMs like \ours{} can generate code, the question remains: how can we efficiently bridge the gap between LLM-generated code and the specific requirements of enterprise data infrastructures? One possible solution lies in the development of encapsulated programming languages, such as Logic-LM~\cite{pan2023logic}, which provide a structured framework that can be tailored to specific use cases.

In previous version of \ours{}~\cite{zha2023tablegptunifyingtablesnature}, we adopted a mixed output approach that combines both structured DSL output and standard programming code. The generation of this hybrid output was guided by prompt templates, and further reinforced during the supervised fine-tuning (SFT) process. This allowed the model to fluidly produce both structured and unstructured code, offering flexibility while maintaining the structure needed for domain-specific applications.

These domain-specific languages offer several benefits, such as better interpretability, and allowing users to interact more directly with the LLM’s output through a user-friendly interface. Additionally, these languages can lead to safer and more robust solutions by minimizing potential security risks and errors. However, this encapsulation can limit the flexibility of the code, as it is constrained within predefined structures.

In conclusion, coding in production environments goes beyond simple code generation. It requires careful consideration of domain-specific needs, infrastructure compatibility, and the ability to strike a balance between flexibility and safety, especially when using mixed approaches of DSL and general-purpose code.

\subsection{Multi-agent Design}

Although \ours{} achieves state-of-the-art performance in table-related tasks, we still cannot yet expect a single end-to-end LLM to fully solve complex, real-world problems independently. Recently, we have been closely following a new line of research focused on automated agency design~\cite{hu2024ADAS}, which builds on the principles to automate the orchestration of LLM workflows. In this approach, several LLMs are organized into a directed acyclic graph (DAG) structure, such that the input queries are automatically routed through a sequence of LLMs by the topological order of the graph. Each LLM performs a specialized functionality. The flow through the DAG is determined by the system itself, making decisions about which LLMs to involve based on the task at hand. These models may vary in their system prompt templates, retrieval-augmented generation (RAG) configurations, in-context learning (ICL) examples, and others. This automated flow engineering creates a flexible, modular pipeline that adjusts dynamically based on the problem’s requirements, much like how AutoML systems automatically configure machine learning models for optimal performance.

For example, in our small app designed for equity and fund recommendations via natural language, we needed to connect the LLM with real-time market data. In this multi-agent architecture, we generally assign distinct roles to different LLMs, each fine-tuned (SFT-ed) on data specifically tailored to its function. One LLM is dedicated to precise intent recognition, trained with datasets focused on understanding user queries and intents. Another LLM is specialized in code generation, data interaction, and tool invocation. A third LLM handles domain-specific and in-depth analysis, fine-tuned on industry-specific datasets to ensure expertise in the relevant field. Each LLM is further configured with distinct prompt templates and retrieval-augmented generation (RAG) setups for the input, while varied coding or conversation logic is applied to the output stage. This tailored fine-tuning at each stage ensures that the overall pipeline delivers precise, accurate, and context-aware responses, addressing the complex nature of real-world applications.

Indeed, while there is ongoing speculation that a single foundation model with sufficiently advanced capabilities might eventually replace the need for chaining multiple models, this remains largely theoretical. Such a model would possess enough general-purpose intelligence to handle various tasks within a unified framework. However, based on our experience with specialized projects, we typically require more than two LLMs to address the full complexity of real-world applications. Just as MetaGPT~\cite{hong2023metagpt} suggests, unified models are promising in their design, but it is evident that for now, one LLM is not sufficient to manage the intricacies of complex workflows. Much like the Qwen and LLaMA model families, where specialized models are developed for tasks such as mathematics or coding, it remains uncertain when a single model can seamlessly solve problems across multiple domains at a high level of proficiency, especially towards production.

\subsection{Tables Are Versatile}
While \ours{} primarily focuses on business intelligence (BI) applications, where databases or data warehouses serve as the upstream sources of structured data, another common and unneglectable form of tabular data originates from apps like Apple Pages or Microsoft Excel. These types of tables differ significantly from those in data infrastructures because they often exhibit irregularities. For example, tables in Excel or Pages frequently have merged cells, inconsistent row or column structures, and non-standard data formatting, making them more complex to process. These tables can vary widely in organization, where cells might contain free-form text, be partially filled, or use multi-level headers, making them far less uniform than typical database tables.

In the agent workflow where \ours{} model resides, we fine-tuned a separate LLM specifically for normalizing irregular tables and integrated it into a holistic system. However, handling these irregular tables still leaves considerable room for improvement, especially given the substantial commercial potential for production use.

Thus, we may hypothesize that addressing such irregularities should begin during the pretraining stage to ensure the model becomes adept at handling the wide array of formats that tables can take. Many current LLMs, along with retrieval-augmented generation (RAG) processes, do not adequately process or handle these non-standard table structures. Moreover, many existing pretraining corpora tend to overlook this type of data~\cite{gunasekar2023textbooksneed}. This gap presents a valuable opportunity for future research to enhance models capable of effectively processing versatile structured data formats, particularly those widely used, such as Excel and Pages.
\section*{Contribution List}

(\textbf{Bolded author name indicates a directional lead.} Listed alphabetically by the first name.)

\paragraph{\textit{Data collection}} Aowen Wang, Hao Chen, \textbf{Kaizhe Shou}, Zhiqing Xiao

\paragraph{\textit{Continual pretraining}} Haokai Xu,  Haoze Li, Qi Zhang, \textbf{Xiaomeng Hu}

\paragraph{\textit{Supervised Fine-tuning}} Ga Zhang,  Jing Yuan, \textbf{Liangyu Zha}, Qingyi Huang, Saisai Yang, 

\paragraph{\textit{Table encoder}} Chao Ye, Haoxuan Lan, Jiaming Tian, \textbf{Lin Long}, \textbf{Liyao Li}, Xijun Gu, Xinjie Sun

\paragraph{\textit{Agent workflow}} Aofeng Su, Chen Zhou, \textbf{Junlin Zhou}, Tao Zhang, Xiang Li

\paragraph{\textit{Benchmark}} Guangcheng Zhu, Pengzuo Wu, Yuhang Yang

\paragraph{\textit{Project support}} Wufang Zhu

\paragraph{\textit{General project lead}} \textbf{Gang Chen}, Haobo Wang, \textbf{Junbo Zhao}, Wentao Ye

\begin{ack}
The project team sincerely thanks the Alibaba Qwen Team for their invaluable support with building \ours{} upon their Qwen-2 and Qwen-2.5 models, with special appreciation to Yiheng Liuda, Junyang Lin, and Chang Zhou.

Project lead Junbo Zhao gratefully acknowledges Zhejiang University’s start-up funding package and the dedicated support from the Information Technology Center in computing power operations and maintenance.
\end{ack}

{
\small
\bibliography{reference}

\begin{thebibliography}{10}

\bibitem{qwenteam2024qwen25}
QwenTeam.
\newblock Qwen2.5, 2024.
\newblock \url{https://qwenlm.github.io/zh/blog/qwen2.5/}.

\bibitem{dubey2024llama}
Abhimanyu Dubey, Abhinav Jauhri, Abhinav Pandey, Abhishek Kadian, Ahmad Al-Dahle, Aiesha Letman, Akhil Mathur, Alan Schelten, Amy Yang, Angela Fan, et~al.
\newblock The llama 3 herd of models.
\newblock {\em arXiv preprint arXiv:2407.21783}, 2024.

\bibitem{guo2024deepseek}
Daya Guo, Qihao Zhu, Dejian Yang, Zhenda Xie, Kai Dong, Wentao Zhang, Guanting Chen, Xiao Bi, Yu~Wu, YK~Li, et~al.
\newblock Deepseek-coder: When the large language model meets programming--the rise of code intelligence.
\newblock {\em arXiv preprint arXiv:2401.14196}, 2024.

\bibitem{pourreza2024chasesqlmultipathreasoningpreference}
Mohammadreza Pourreza, Hailong Li, Ruoxi Sun, Yeounoh Chung, Shayan Talaei, Gaurav~Tarlok Kakkar, Yu~Gan, Amin Saberi, Fatma Ozcan, and Sercan~O. Arik.
\newblock Chase-sql: Multi-path reasoning and preference optimized candidate selection in text-to-sql, 2024.

\bibitem{openai2023chatgpt}
OpenAI.
\newblock Chatgpt, 2022.
\newblock \url{https://openai.com/blog/chatgpt}.

\bibitem{zhong2017seq2sql}
Victor Zhong, Caiming Xiong, and Richard Socher.
\newblock Seq2sql: Generating structured queries from natural language using reinforcement learning.
\newblock {\em arXiv preprint arXiv:1709.00103}, 2017.

\bibitem{achiam2023gpt}
Josh Achiam, Steven Adler, Sandhini Agarwal, Lama Ahmad, Ilge Akkaya, Florencia~Leoni Aleman, Diogo Almeida, Janko Altenschmidt, Sam Altman, Shyamal Anadkat, et~al.
\newblock Gpt-4 technical report.
\newblock {\em arXiv preprint arXiv:2303.08774}, 2023.

\bibitem{zhang2024vision}
Jingyi Zhang, Jiaxing Huang, Sheng Jin, and Shijian Lu.
\newblock Vision-language models for vision tasks: A survey.
\newblock {\em IEEE Transactions on Pattern Analysis and Machine Intelligence}, 2024.

\bibitem{radford2021learning}
Alec Radford, Jong~Wook Kim, Chris Hallacy, Aditya Ramesh, Gabriel Goh, Sandhini Agarwal, Girish Sastry, Amanda Askell, Pamela Mishkin, Jack Clark, et~al.
\newblock Learning transferable visual models from natural language supervision.
\newblock In {\em International conference on machine learning}, pages 8748--8763. PMLR, 2021.

\bibitem{ramesh2021zeroshottexttoimagegeneration}
Aditya Ramesh, Mikhail Pavlov, Gabriel Goh, Scott Gray, Chelsea Voss, Alec Radford, Mark Chen, and Ilya Sutskever.
\newblock Zero-shot text-to-image generation, 2021.

\bibitem{li2024personal}
Yuanchun Li, Hao Wen, Weijun Wang, Xiangyu Li, Yizhen Yuan, Guohong Liu, Jiacheng Liu, Wenxing Xu, Xiang Wang, Yi~Sun, et~al.
\newblock Personal llm agents: Insights and survey about the capability, efficiency and security.
\newblock {\em arXiv preprint arXiv:2401.05459}, 2024.

\bibitem{fang2024large}
Xi~Fang, Weijie Xu, Fiona~Anting Tan, Jiani Zhang, Ziqing Hu, Yanjun~Jane Qi, Scott Nickleach, Diego Socolinsky, Srinivasan Sengamedu, Christos Faloutsos, et~al.
\newblock Large language models (llms) on tabular data: Prediction, generation, and understanding-a survey.
\newblock 2024.

\bibitem{hu2023chatdb}
Chenxu Hu, Jie Fu, Chenzhuang Du, Simian Luo, Junbo Zhao, and Hang Zhao.
\newblock Chatdb: Augmenting llms with databases as their symbolic memory.
\newblock {\em arXiv preprint arXiv:2306.03901}, 2023.

\bibitem{li2023graphix}
Jinyang Li, Binyuan Hui, Reynold Cheng, Bowen Qin, Chenhao Ma, Nan Huo, Fei Huang, Wenyu Du, Luo Si, and Yongbin Li.
\newblock Graphix-t5: Mixing pre-trained transformers with graph-aware layers for text-to-sql parsing.
\newblock In {\em Proceedings of the AAAI Conference on Artificial Intelligence}, volume~37, pages 13076--13084, 2023.

\bibitem{jiang2022omnitab}
Zhengbao Jiang, Yi~Mao, Pengcheng He, Graham Neubig, and Weizhu Chen.
\newblock Omnitab: Pretraining with natural and synthetic data for few-shot table-based question answering.
\newblock In {\em Proceedings of the 2022 Conference of the North American Chapter of the Association for Computational Linguistics: Human Language Technologies}, pages 932--942, 2022.

\bibitem{ye2023large}
Yunhu Ye, Binyuan Hui, Min Yang, Binhua Li, Fei Huang, and Yongbin Li.
\newblock Large language models are versatile decomposers: Decomposing evidence and questions for table-based reasoning.
\newblock In {\em Proceedings of the 46th International ACM SIGIR Conference on Research and Development in Information Retrieval}, pages 174--184, 2023.

\bibitem{zhang2024tablellama}
Tianshu Zhang, Xiang Yue, Yifei Li, and Huan Sun.
\newblock Tablellama: Towards open large generalist models for tables.
\newblock In {\em Proceedings of the 2024 Conference of the North American Chapter of the Association for Computational Linguistics: Human Language Technologies (Volume 1: Long Papers)}, pages 6024--6044, 2024.

\bibitem{zha2023tablegptunifyingtablesnature}
Liangyu Zha, Junlin Zhou, Liyao Li, Rui Wang, Qingyi Huang, Saisai Yang, Jing Yuan, Changbao Su, Xiang Li, Aofeng Su, Tao Zhang, Chen Zhou, Kaizhe Shou, Miao Wang, Wufang Zhu, Guoshan Lu, Chao Ye, Yali Ye, Wentao Ye, Yiming Zhang, Xinglong Deng, Jie Xu, Haobo Wang, Gang Chen, and Junbo Zhao.
\newblock Tablegpt: Towards unifying tables, nature language and commands into one gpt, 2023.

\bibitem{ke2023continual}
Zixuan Ke, Yijia Shao, Haowei Lin, Tatsuya Konishi, Gyuhak Kim, and Bing Liu.
\newblock Continual pre-training of language models.
\newblock {\em arXiv preprint arXiv:2302.03241}, 2023.

\bibitem{liu2024deepseek}
Aixin Liu, Bei Feng, Bin Wang, Bingxuan Wang, Bo~Liu, Chenggang Zhao, Chengqi Dengr, Chong Ruan, Damai Dai, Daya Guo, et~al.
\newblock Deepseek-v2: A strong, economical, and efficient mixture-of-experts language model.
\newblock {\em arXiv preprint arXiv:2405.04434}, 2024.

\bibitem{lin2024rho}
Zhenghao Lin, Zhibin Gou, Yeyun Gong, Xiao Liu, Yelong Shen, Ruochen Xu, Chen Lin, Yujiu Yang, Jian Jiao, Nan Duan, et~al.
\newblock Rho-1: Not all tokens are what you need.
\newblock {\em arXiv preprint arXiv:2404.07965}, 2024.

\bibitem{kim2024strategicdataorderingenhancing}
Jisu Kim and Juhwan Lee.
\newblock Strategic data ordering: Enhancing large language model performance through curriculum learning, 2024.

\bibitem{hurst2024gpt}
Aaron Hurst, Adam Lerer, Adam~P Goucher, Adam Perelman, Aditya Ramesh, Aidan Clark, AJ~Ostrow, Akila Welihinda, Alan Hayes, Alec Radford, et~al.
\newblock Gpt-4o system card.
\newblock {\em arXiv preprint arXiv:2410.21276}, 2024.

\bibitem{li2023blip}
Junnan Li, Dongxu Li, Silvio Savarese, and Steven Hoi.
\newblock Blip-2: Bootstrapping language-image pre-training with frozen image encoders and large language models.
\newblock In {\em International conference on machine learning}, pages 19730--19742. PMLR, 2023.

\bibitem{lewis2020retrieval}
Patrick Lewis, Ethan Perez, Aleksandra Piktus, Fabio Petroni, Vladimir Karpukhin, Naman Goyal, Heinrich K{\"u}ttler, Mike Lewis, Wen-tau Yih, Tim Rockt{\"a}schel, et~al.
\newblock Retrieval-augmented generation for knowledge-intensive nlp tasks.
\newblock {\em Advances in Neural Information Processing Systems}, 33:9459--9474, 2020.

\bibitem{kaggle}
Kaggle.
\newblock Kaggle datasets.
\newblock \url{https://www.kaggle.com/datasets}.

\bibitem{dodge2021documentinglargewebtextcorpora}
Jesse Dodge, Maarten Sap, Ana Marasović, William Agnew, Gabriel Ilharco, Dirk Groeneveld, Margaret Mitchell, and Matt Gardner.
\newblock Documenting large webtext corpora: A case study on the colossal clean crawled corpus, 2021.

\bibitem{penedo2023refinedwebdatasetfalconllm}
Guilherme Penedo, Quentin Malartic, Daniel Hesslow, Ruxandra Cojocaru, Alessandro Cappelli, Hamza Alobeidli, Baptiste Pannier, Ebtesam Almazrouei, and Julien Launay.
\newblock The refinedweb dataset for falcon llm: Outperforming curated corpora with web data, and web data only, 2023.

\bibitem{lozhkov2024starcoder2stackv2}
Anton Lozhkov, Raymond Li, Loubna~Ben Allal, Federico Cassano, Joel Lamy-Poirier, Nouamane Tazi, Ao~Tang, Dmytro Pykhtar, Jiawei Liu, Yuxiang Wei, Tianyang Liu, Max Tian, Denis Kocetkov, Arthur Zucker, Younes Belkada, Zijian Wang, Qian Liu, Dmitry Abulkhanov, Indraneil Paul, Zhuang Li, Wen-Ding Li, Megan Risdal, Jia Li, Jian Zhu, Terry~Yue Zhuo, Evgenii Zheltonozhskii, Nii Osae~Osae Dade, Wenhao Yu, Lucas Krauß, Naman Jain, Yixuan Su, Xuanli He, Manan Dey, Edoardo Abati, Yekun Chai, Niklas Muennighoff, Xiangru Tang, Muhtasham Oblokulov, Christopher Akiki, Marc Marone, Chenghao Mou, Mayank Mishra, Alex Gu, Binyuan Hui, Tri Dao, Armel Zebaze, Olivier Dehaene, Nicolas Patry, Canwen Xu, Julian McAuley, Han Hu, Torsten Scholak, Sebastien Paquet, Jennifer Robinson, Carolyn~Jane Anderson, Nicolas Chapados, Mostofa Patwary, Nima Tajbakhsh, Yacine Jernite, Carlos~Muñoz Ferrandis, Lingming Zhang, Sean Hughes, Thomas Wolf, Arjun Guha, Leandro von Werra, and Harm de~Vries.
\newblock Starcoder 2 and the stack v2: The next generation, 2024.

\bibitem{joulin2016bagtricksefficienttext}
Armand Joulin, Edouard Grave, Piotr Bojanowski, and Tomas Mikolov.
\newblock Bag of tricks for efficient text classification, 2016.

\bibitem{li2024codes}
Haoyang Li, Jing Zhang, Hanbing Liu, Ju~Fan, Xiaokang Zhang, Jun Zhu, Renjie Wei, Hongyan Pan, Cuiping Li, and Hong Chen.
\newblock Codes: Towards building open-source language models for text-to-sql.
\newblock {\em Proceedings of the ACM on Management of Data}, 2(3):1--28, 2024.

\bibitem{pourreza2024din}
Mohammadreza Pourreza and Davood Rafiei.
\newblock Din-sql: Decomposed in-context learning of text-to-sql with self-correction.
\newblock {\em Advances in Neural Information Processing Systems}, 36, 2024.

\bibitem{li2024unifying}
Shujie Li, Liang Li, Ruiying Geng, Min Yang, Binhua Li, Guanghu Yuan, Wanwei He, Shao Yuan, Can Ma, Fei Huang, et~al.
\newblock Unifying structured data as graph for data-to-text pre-training.
\newblock {\em Transactions of the Association for Computational Linguistics}, 12:210--228, 2024.

\bibitem{yang2022tableformer}
Jingfeng Yang, Aditya Gupta, Shyam Upadhyay, Luheng He, Rahul Goel, and Shachi Paul.
\newblock Tableformer: Robust transformer modeling for table-text encoding.
\newblock {\em arXiv preprint arXiv:2203.00274}, 2022.

\bibitem{yin2020tabert}
Pengcheng Yin, Graham Neubig, Wen-tau Yih, and Sebastian Riedel.
\newblock Tabert: Pretraining for joint understanding of textual and tabular data.
\newblock {\em arXiv preprint arXiv:2005.08314}, 2020.

\bibitem{somepalli2021saint}
Gowthami Somepalli, Micah Goldblum, Avi Schwarzschild, C~Bayan Bruss, and Tom Goldstein.
\newblock Saint: Improved neural networks for tabular data via row attention and contrastive pre-training.
\newblock {\em arXiv preprint arXiv:2106.01342}, 2021.

\bibitem{zhu2023xtab}
Bingzhao Zhu, Xingjian Shi, Nick Erickson, Mu~Li, George Karypis, and Mahsa Shoaran.
\newblock Xtab: Cross-table pretraining for tabular transformers.
\newblock {\em arXiv preprint arXiv:2305.06090}, 2023.

\bibitem{kenton2019bert}
Jacob Devlin Ming-Wei~Chang Kenton and Lee~Kristina Toutanova.
\newblock Bert: Pre-training of deep bidirectional transformers for language understanding.
\newblock In {\em Proceedings of naacL-HLT}, volume~1, page~2. Minneapolis, Minnesota, 2019.

\bibitem{bai2023qwen}
Jinze Bai, Shuai Bai, Shusheng Yang, Shijie Wang, Sinan Tan, Peng Wang, Junyang Lin, Chang Zhou, and Jingren Zhou.
\newblock Qwen-vl: A frontier large vision-language model with versatile abilities.
\newblock {\em arXiv preprint arXiv:2308.12966}, 2023.

\bibitem{wang2024qwen2}
Peng Wang, Shuai Bai, Sinan Tan, Shijie Wang, Zhihao Fan, Jinze Bai, Keqin Chen, Xuejing Liu, Jialin Wang, Wenbin Ge, et~al.
\newblock Qwen2-vl: Enhancing vision-language model's perception of the world at any resolution.
\newblock {\em arXiv preprint arXiv:2409.12191}, 2024.

\bibitem{he2020momentum}
Kaiming He, Haoqi Fan, Yuxin Wu, Saining Xie, and Ross Girshick.
\newblock Momentum contrast for unsupervised visual representation learning.
\newblock In {\em Proceedings of the IEEE/CVF conference on computer vision and pattern recognition}, pages 9729--9738, 2020.

\bibitem{chen2020simple}
Ting Chen, Simon Kornblith, Mohammad Norouzi, and Geoffrey Hinton.
\newblock A simple framework for contrastive learning of visual representations.
\newblock In {\em International conference on machine learning}, pages 1597--1607. PMLR, 2020.

\bibitem{Nan2021FeTaQAFT}
Linyong Nan, Chiachun Hsieh, Ziming Mao, Xi~Victoria Lin, Neha Verma, Rui Zhang, Wojciech Kryściński, Hailey Schoelkopf, Riley Kong, Xiangru Tang, Mutethia Mutuma, Ben Rosand, Isabel Trindade, Renusree Bandaru, Jacob Cunningham, Caiming Xiong, and Dragomir Radev.
\newblock Fetaqa: Free-form table question answering.
\newblock {\em Transactions of the Association for Computational Linguistics}, 10:35--49, 2022.

\bibitem{wikitq}
Panupong Pasupat and Percy Liang.
\newblock Compositional semantic parsing on semi-structured tables.
\newblock In {\em Proceedings of the 53rd Annual Meeting of the Association for Computational Linguistics and the 7th International Joint Conference on Natural Language Processing (Volume 1: Long Papers)}, pages 1470--1480, Beijing, China, July 2015. Association for Computational Linguistics.

\bibitem{parikh-etal-2020-totto}
Ankur Parikh, Xuezhi Wang, Sebastian Gehrmann, Manaal Faruqui, Bhuwan Dhingra, Diyi Yang, and Dipanjan Das.
\newblock {ToTTo}: A controlled table-to-text generation dataset.
\newblock In {\em Proceedings of the 2020 Conference on Empirical Methods in Natural Language Processing (EMNLP)}, pages 1173--1186, Online, November 2020. Association for Computational Linguistics.

\bibitem{anonymous2024rethinking}
Anonymous.
\newblock Rethinking table instruction tuning.
\newblock In {\em Submitted to The Thirteenth International Conference on Learning Representations}, 2024.
\newblock under review.

\bibitem{uci}
UC~Irvine.
\newblock Uci machine learning repository.
\newblock \url{http://archive.ics.uci.edu/datasets}.

\bibitem{tianchi}
Aliyun.
\newblock Tianchi datasets.
\newblock \url{https://tianchi.aliyun.com/dataset}.

\bibitem{glm2024chatglm}
Team GLM, Aohan Zeng, Bin Xu, Bowen Wang, Chenhui Zhang, Da~Yin, Diego Rojas, Guanyu Feng, Hanlin Zhao, Hanyu Lai, et~al.
\newblock Chatglm: A family of large language models from glm-130b to glm-4 all tools.
\newblock {\em arXiv preprint arXiv:2406.12793}, 2024.

\bibitem{yicoder}
01.AI.
\newblock Meet yi-coder: A small but mighty llm for code, September 2024.

\bibitem{hui2024qwen25codertechnicalreport}
Binyuan Hui, Jian Yang, Zeyu Cui, Jiaxi Yang, Dayiheng Liu, Lei Zhang, Tianyu Liu, Jiajun Zhang, Bowen Yu, Kai Dang, An~Yang, Rui Men, Fei Huang, Xingzhang Ren, Xuancheng Ren, Jingren Zhou, and Junyang Lin.
\newblock Qwen2.5-coder technical report, 2024.

\bibitem{zhang2024tablellm}
Xiaokang Zhang, Jing Zhang, Zeyao Ma, Yang Li, Bohan Zhang, Guanlin Li, Zijun Yao, Kangli Xu, Jinchang Zhou, Daniel Zhang-Li, et~al.
\newblock Tablellm: Enabling tabular data manipulation by llms in real office usage scenarios.
\newblock {\em arXiv preprint arXiv:2403.19318}, 2024.

\bibitem{wu2024tablebench}
Xianjie Wu, Jian Yang, Linzheng Chai, Ge~Zhang, Jiaheng Liu, Xinrun Du, Di~Liang, Daixin Shu, Xianfu Cheng, Tianzhen Sun, et~al.
\newblock Tablebench: A comprehensive and complex benchmark for table question answering.
\newblock {\em arXiv preprint arXiv:2408.09174}, 2024.

\bibitem{roziere2023code}
Baptiste Roziere, Jonas Gehring, Fabian Gloeckle, Sten Sootla, Itai Gat, Xiaoqing~Ellen Tan, Yossi Adi, Jingyu Liu, Romain Sauvestre, Tal Remez, et~al.
\newblock Code llama: Open foundation models for code.
\newblock {\em arXiv preprint arXiv:2308.12950}, 2023.

\bibitem{yang2024qwen2}
An~Yang, Baosong Yang, Binyuan Hui, Bo~Zheng, Bowen Yu, Chang Zhou, Chengpeng Li, Chengyuan Li, Dayiheng Liu, Fei Huang, et~al.
\newblock Qwen2 technical report.
\newblock {\em arXiv preprint arXiv:2407.10671}, 2024.

\bibitem{qwen}
Jinze Bai, Shuai Bai, Yunfei Chu, Zeyu Cui, Kai Dang, Xiaodong Deng, Yang Fan, Wenbin Ge, Yu~Han, Fei Huang, Binyuan Hui, Luo Ji, Mei Li, Junyang Lin, Runji Lin, Dayiheng Liu, Gao Liu, Chengqiang Lu, Keming Lu, Jianxin Ma, Rui Men, Xingzhang Ren, Xuancheng Ren, Chuanqi Tan, Sinan Tan, Jianhong Tu, Peng Wang, Shijie Wang, Wei Wang, Shengguang Wu, Benfeng Xu, Jin Xu, An~Yang, Hao Yang, Jian Yang, Shusheng Yang, Yang Yao, Bowen Yu, Hongyi Yuan, Zheng Yuan, Jianwei Zhang, Xingxuan Zhang, Yichang Zhang, Zhenru Zhang, Chang Zhou, Jingren Zhou, Xiaohuan Zhou, and Tianhang Zhu.
\newblock Qwen technical report.
\newblock {\em arXiv preprint arXiv:2309.16609}, 2023.

\bibitem{lee2024mcs}
Dongjun Lee, Choongwon Park, Jaehyuk Kim, and Heesoo Park.
\newblock Mcs-sql: Leveraging multiple prompts and multiple-choice selection for text-to-sql generation.
\newblock {\em arXiv preprint arXiv:2405.07467}, 2024.

\bibitem{deng2020turl}
Xiang Deng, Huan Sun, Alyssa Lees, You Wu, and Cong Yu.
\newblock Turl: table understanding through representation learning.
\newblock {\em Proceedings of the VLDB Endowment}, 14(3):307--319, 2020.

\bibitem{wikisql}
Victor Zhong, Caiming Xiong, and Richard Socher.
\newblock Seq2sql: Generating structured queries from natural language using reinforcement learning.
\newblock {\em CoRR}, abs/1709.00103, 2017.

\bibitem{cheng-etal-2022-hitab}
Zhoujun Cheng, Haoyu Dong, Zhiruo Wang, Ran Jia, Jiaqi Guo, Yan Gao, Shi Han, Jian-Guang Lou, and Dongmei Zhang.
\newblock {H}i{T}ab: A hierarchical table dataset for question answering and natural language generation.
\newblock In {\em Proceedings of the 60th Annual Meeting of the Association for Computational Linguistics (Volume 1: Long Papers)}, pages 1094--1110, Dublin, Ireland, May 2022. Association for Computational Linguistics.

\bibitem{chen2020hybridqa}
Wenhu Chen, Hanwen Zha, Zhiyu Chen, Wenhan Xiong, Hong Wang, and William~Yang Wang.
\newblock Hybridqa: A dataset of multi-hop question answering over tabular and textual data.
\newblock In {\em Findings of the Association for Computational Linguistics: EMNLP 2020}, pages 1026--1036, 2020.

\bibitem{Chen2020TabFact}
Wenhu Chen, Hongmin Wang, Jianshu Chen, Yunkai Zhang, Hong Wang, Shiyang Li, Xiyou Zhou, and William~Yang Wang.
\newblock Tabfact: A large-scale dataset for table-based fact verification.
\newblock In {\em International Conference on Learning Representations}, 2020.

\bibitem{feverous}
Rami Aly, Zhijiang Guo, Michael~Sejr Schlichtkrull, James Thorne, Andreas Vlachos, Christos Christodoulopoulos, Oana Cocarascu, and Arpit Mittal.
\newblock The fact extraction and {VER}ification over unstructured and structured information ({FEVEROUS}) shared task.
\newblock In {\em Proceedings of the Fourth Workshop on Fact Extraction and VERification (FEVER)}, pages 1--13, Dominican Republic, November 2021. Association for Computational Linguistics.

\bibitem{li2024bird}
Jinyang Li, Binyuan Hui, Ge~Qu, Jiaxi Yang, Binhua Li, Bowen Li, Bailin Wang, Bowen Qin, Ruiying Geng, Nan Huo, et~al.
\newblock Can llm already serve as a database interface? a big bench for large-scale database grounded text-to-sqls.
\newblock {\em Advances in Neural Information Processing Systems}, 36, 2024.

\bibitem{yu2018spider}
Tao Yu, Rui Zhang, Kai Yang, Michihiro Yasunaga, Dongxu Wang, Zifan Li, James Ma, Irene Li, Qingning Yao, Shanelle Roman, et~al.
\newblock Spider: A large-scale human-labeled dataset for complex and cross-domain semantic parsing and text-to-sql task.
\newblock In {\em Proceedings of the 2018 Conference on Empirical Methods in Natural Language Processing}, pages 3911--3921, 2018.

\bibitem{chen2021evaluating}
Mark Chen, Jerry Tworek, Heewoo Jun, Qiming Yuan, Henrique Ponde De~Oliveira Pinto, Jared Kaplan, Harri Edwards, Yuri Burda, Nicholas Joseph, Greg Brockman, et~al.
\newblock Evaluating large language models trained on code.
\newblock {\em arXiv preprint arXiv:2107.03374}, 2021.

\bibitem{austin2021program}
Jacob Austin, Augustus Odena, Maxwell Nye, Maarten Bosma, Henryk Michalewski, David Dohan, Ellen Jiang, Carrie Cai, Michael Terry, Quoc Le, et~al.
\newblock Program synthesis with large language models.
\newblock {\em arXiv preprint arXiv:2108.07732}, 2021.

\bibitem{li2023cmmlu}
Haonan Li, Yixuan Zhang, Fajri Koto, Yifei Yang, Hai Zhao, Yeyun Gong, Nan Duan, and Timothy Baldwin.
\newblock Cmmlu: Measuring massive multitask language understanding in chinese.
\newblock {\em arXiv preprint arXiv:2306.09212}, 2023.

\bibitem{hendrycks2020measuring}
Dan Hendrycks, Collin Burns, Steven Basart, Andy Zou, Mantas Mazeika, Dawn Song, and Jacob Steinhardt.
\newblock Measuring massive multitask language understanding.
\newblock {\em arXiv preprint arXiv:2009.03300}, 2020.

\bibitem{pan2023logic}
Liangming Pan, Alon Albalak, Xinyi Wang, and William~Yang Wang.
\newblock Logic-lm: Empowering large language models with symbolic solvers for faithful logical reasoning.
\newblock {\em arXiv preprint arXiv:2305.12295}, 2023.

\bibitem{hu2024ADAS}
Shengran Hu, Cong Lu, and Jeff Clune.
\newblock Automated design of agentic systems.
\newblock {\em arXiv preprint arXiv:2408.08435}, 2024.

\bibitem{hong2023metagpt}
Sirui Hong, Xiawu Zheng, Jonathan Chen, Yuheng Cheng, Jinlin Wang, Ceyao Zhang, Zili Wang, Steven Ka~Shing Yau, Zijuan Lin, Liyang Zhou, et~al.
\newblock Metagpt: Meta programming for multi-agent collaborative framework.
\newblock {\em arXiv preprint arXiv:2308.00352}, 2023.

\bibitem{gunasekar2023textbooksneed}
Suriya Gunasekar, Yi~Zhang, Jyoti Aneja, Caio César~Teodoro Mendes, Allie~Del Giorno, Sivakanth Gopi, Mojan Javaheripi, Piero Kauffmann, Gustavo de~Rosa, Olli Saarikivi, Adil Salim, Shital Shah, Harkirat~Singh Behl, Xin Wang, Sébastien Bubeck, Ronen Eldan, Adam~Tauman Kalai, Yin~Tat Lee, and Yuanzhi Li.
\newblock Textbooks are all you need, 2023.

\end{thebibliography}
\bibliographystyle{unsrt}
}

\end{document}